\pdfoutput=1

\documentclass[11pt]{article}

\usepackage{acl}
\usepackage[T1]{fontenc}
\usepackage{CJKutf8}

\usepackage{microtype}

\usepackage{inconsolata}

\usepackage{multirow}
\usepackage[T5]{fontenc}
\usepackage{bm}
\usepackage{listings}
\usepackage{tabularx}
\usepackage{ltablex}
\usepackage{longtable}
\usepackage{graphicx}
\usepackage{colortbl}

\usepackage{pythonhighlight}

\usepackage{supertabular}
\usepackage{xltabular}

\usepackage{adjustbox}
\usepackage{lineno}
\usepackage{array}
\usepackage{booktabs} 
\usepackage{hyperref}
\usepackage{amsmath}
\usepackage{amsfonts}
\usepackage{amssymb}
\usepackage{makecell}
\usepackage{paralist}

\title{wav2graph: A Framework for Supervised Learning \\Knowledge Graph from Speech}

\author{Khai Le-Duc$^{*1,2,4}$, Quy-Anh Dang$^{*3,6}$, Tan-Hanh Pham$^5$, 
Truong-Son Hy$^{4,7}$\\
$^1$University of Toronto, Canada
$^2$University Health Network, Canada \\
$^3$VNU University of Science, Vietnam
$^4$FPT Software AI Center, Vietnam\\
$^5$Florida Institute of Technology, USA \\
$^6$Knovel Engineering Lab, Singapore
$^7$University of Alabama at Birmingham, USA\\
  \texttt{duckhai.le@mail.utoronto.ca, thy@uab.edu}
  \\}

\begin{document}
\maketitle
\begin{abstract}
Knowledge graphs (KGs) enhance the performance of large language models (LLMs) and search engines by providing structured, interconnected data that improves reasoning and context-awareness. However, KGs only focus on text data, thereby neglecting other modalities such as speech. In this work, we introduce \textit{wav2graph}, the first framework for supervised learning knowledge graph from speech data. Our pipeline are straightforward: (1) constructing a KG based on transcribed spoken utterances and a named entity database, (2) converting KG into embedding vectors, and (3) training graph neural networks (GNNs) for node classification and link prediction tasks. Through extensive experiments conducted in inductive and transductive learning contexts using state-of-the-art GNN models, we provide baseline results and error analysis for node classification and link prediction tasks on human transcripts and automatic speech recognition (ASR) transcripts, including evaluations using both encoder-based and decoder-based node embeddings, as well as monolingual and multilingual acoustic pre-trained models. All related code, data, and models are published online.
\end{abstract}

\def\thefootnote{(*)}\footnotetext{Equal contribution}\def\thefootnote{\arabic{footnote}}

\thispagestyle{plain}
\pagestyle{plain}

\begin{figure*}[h]
    \centering
    \includegraphics[width=\textwidth]{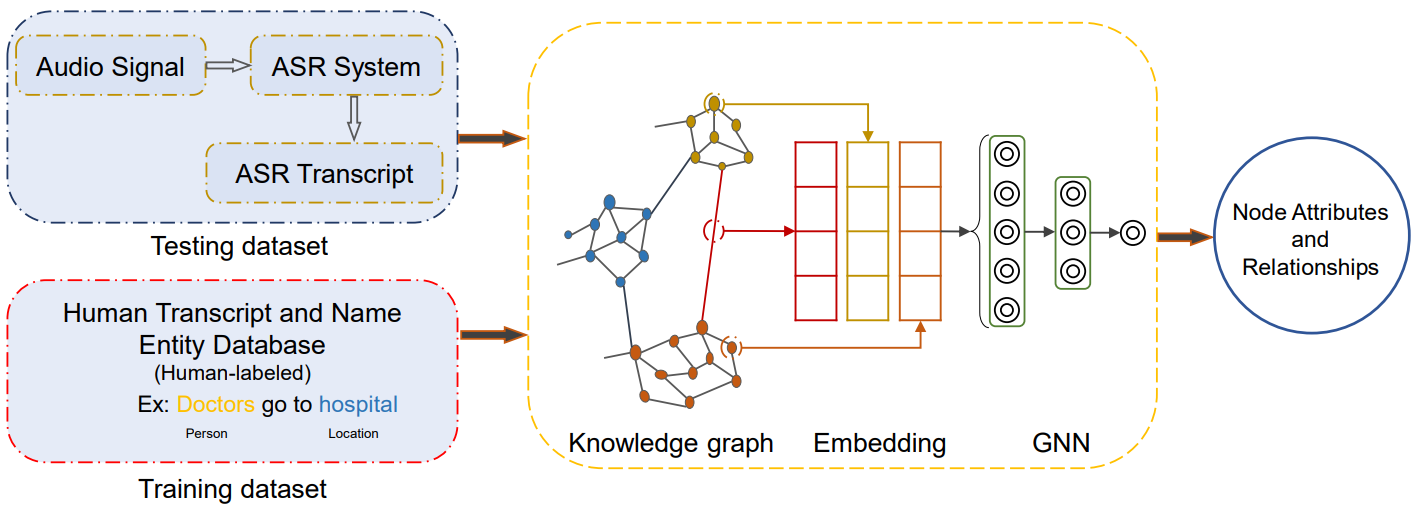}
    \caption{Visualization of our \textit{wav2graph} framework. We train GNNs on the KG that is built from human transcript and its corresponding NEs. Then we infer directly on another KG that is built from ASR transcript to acquire node attributes and node relationships.}
    \label{fig:wav2graph}
\end{figure*}

\section{Introduction}
In the field of Artificial Intelligence (AI), KGs have emerged as a powerful approach to knowledge representation and reasoning. KGs leverage graph-structured models to encode entities (objects, events, concepts) and the relationships linking them \cite{fensel2020introduction, ji2021survey}. This structured representation enables efficient storage, retrieval, and reasoning over vast amounts of interconnected information \cite{hogan2021knowledge, chen2020review}.

The utility of KGs spans a variety of high-impact applications. For example, prominent search engines such as Bing, Google, and Yahoo employ KGs to enhance search relevance and personalization \cite{steiner2012adding, uyar2015evaluating, juel2013ethics}. Knowledge engines and question-answering systems, such as Wolfram Alpha, Siri, and Alexa, leverage KGs to deliver precise and contextually appropriate responses \cite{he2020bert, fei2021enriching}. Social networks, including LinkedIn and Facebook, also utilize KGs to enrich user profiles and enable sophisticated social recommendations \cite{pellissier2016freebase, lehmann2015dbpedia}. Furthermore, over the past three years, KGs have become pivotal in enhancing the reasoning capabilities of LLMs by providing structured, interconnected data that enhances the model's ability to understand and generate contextually relevant and accurate information \cite{pan2024unifying, yasunaga2021qa, ji2020language}.

Despite their significant advantages, the construction and training of voice-based KGs remains a complex and largely unexplored process. Among limited number of relevant works we found, to the best of our knowledge, \citet{fu2021speech} claimed to present the first automatic KG construction system from speech. \citet{wu2022towards} proposed a new information extraction task, speech relation extraction, that used extracted relations between synthetic ASR transcripts to build a KG. However, there has been no training conducted on such KGs to the best of our knowledge. Training KGs is necessary because GNN models can learn to extract and generalize complex patterns and relationships from the data in a KG, enabling the prediction on unseen data in the KG, which direct use of the KG alone (rule-based) cannot achieve.

To address this gap, this paper introduces \textit{wav2graph}, a pioneering framework designed to train KGs directly from speech data. \textit{wav2graph} leverages supervised learning GNNs to automate the process of extracting entities and relationships from spoken language, paving the way for the integration of speech-derived knowledge into various AI applications. Our contribution is summarized as follows:
\begin{itemize}
    \item We propose, to the best of our knowledge, the first framework for supervised learning KGs from speech
    \item We release the first real-world KG from speech
    \item We present empirical baselines and conduct a comprehensive analysis of transductive and inductive learning on both human transcripts and ASR transcripts using state-of-the-art GNNs.  
\end{itemize}

All code, data, and models are publicly available online\footnote{https://github.com/leduckhai/wav2graph}.

\section{Data}
\subsection{Data Collection}
We selected the \textit{VietMed-NER} dataset \cite{leduc2024medical} as an initialization for our KG construction due to its status as the largest spoken named entity recognition (NER) dataset in the world in terms of the number of entity types, characterized by 18 distinct entity types. The dataset focused on real-world medical conversations.

\begin{figure}
    \centering
    \includegraphics[width=\columnwidth]{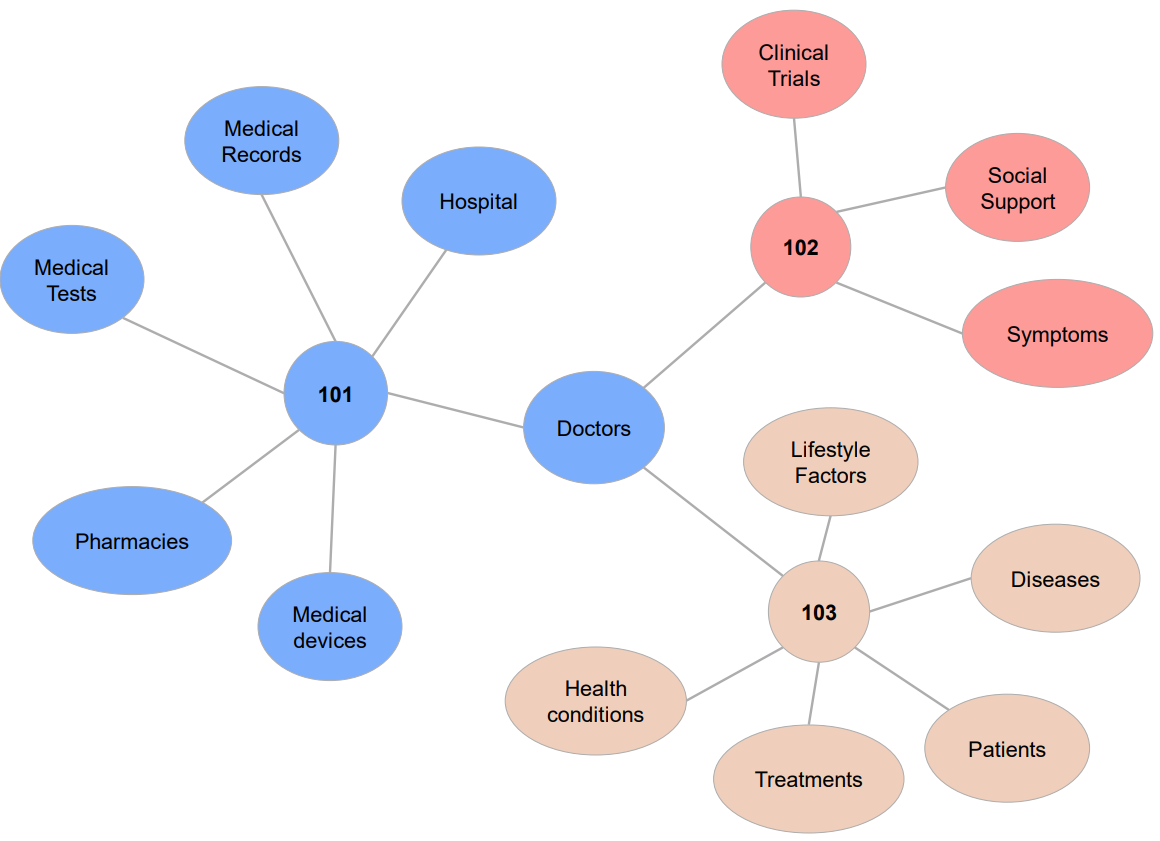}
    \caption{An example of our KG. Node \textit{101}, \textit{102}, \textit{103} is the utterance identification number, while remaining nodes are NEs. We follow entity-utterance-entity approach \cite{al2020named} to present relationship between nodes in the KG.}
    \label{fig:KG_example}
\end{figure}

As shown in Figure \ref{fig:KG_example}, to construct the knowledge graph, we employ an entity-utterance-entity methodology \cite{al2020named}. NER is used to extract named entities (NEs) from text and categorize them into types such as person, location, and organization. Instead of utilizing an automatic NER system based on machine learning like most previous works \cite{thukral2023knowledge, jia2018practical, nie2021ka}, we opted for gold-standard labels provided by human annotators to extract NEs. These NEs are then linked to the corresponding utterances that mention them, forming relational edges. Consequently, our knowledge graph comprises two types of nodes (\textit{entity\_type} attribute): \textit{utterance} and \textit{named\_entity}. Named entities have an additional attribute, as they are classified into 18 distinct types, such as PERSON and LOCATION, etc,.
 
\subsection{Data Statistics}
\begin{table}[ht]
\caption{Data statistics of our knowledge graph}
\resizebox{\columnwidth}{!}{%
\begin{tabular}{lcccc}
\toprule
 & Train & Dev & Test & Total \\ 
 \midrule
\#Nodes & 7228 & 2409 & 2409 & 12046 \\ 
\#Edges & 12782 & 4260 & 4260 & 21302 \\ 
\#utterance nodes & 5523 & 1884 & 1857 & 9264 \\ 
\#named\_entity nodes & 1705 & 525 & 552 & 2782 \\ 
\bottomrule
\end{tabular}%
}

\label{tab:data_stats}
\end{table}

Table \ref{tab:data_stats} shows the data statistics for our knowledge graph.

\section{\textit{wav2graph}}
Let $x^{T}_{1} := x_{1}, x_{2}, ..., x_{T}$ be an audio signal of length $T$ and let $k \in K$ be an NE in the set of all NEs in database, the aim is to build a learning model $f$ that conduct 2 single tasks:
\begin{itemize}
    \item Node classification: Estimating the node attribute probability $p(c|x^{T}_{1} \vee k)$ for each $c \in C$, where $C$ is the number of attribute classes
    \item Link prediction: Estimating the edge probability $p(e|x^{T}_{1}, k)$, where $e \in \{0, 1\}$ is the edge presence between 2 nodes, $k \in K$ is an NE in the set of all NEs
\end{itemize}
Therefore, the decision rule to predict a single class for node classification task is: 
\begin{equation}
x^{T}_{1} \vee k  \to \hat{c} = \arg\max_{c \in C} f(c|x^{T}_{1} \vee k)
\end{equation}
And the decision rule to predict a single class for link prediction task is: 
\begin{equation}
x^{T}_{1}, k  \to \hat{e} = \arg\max_{e \in \{0, 1\}} f(e|x^{T}_{1}, k)
\end{equation}

\subsection{ASR Model}
An ASR model is used to transcribe audio signal into text by mapping an audio signal $x^{T}_{1}$ of length $T$ to the most likely word sequence $w^{N}_{1}$ of length $N$. 
The relation $w^{*}$ between the acoustic and word sequence is:
\begin{equation}
w^{*} = \operatorname{arg}\max_{w_1^N} \, p(w_{1}^{N}|x_{1}^{T})    
\end{equation}

According to Bayes' Theorem, the probability $p(x_{1}^{T})$ can be ignored during maximization because it only serves as a normalization factor and does not affect the outcome:
\begin{equation}
    \begin{split}
     p(w_{1}^{N}|x_{1}^{T}) &= \frac{p(x_{1}^{T}|w_{1}^{N})p(w_{1}^{N})}{p(x_{1}^{T})} \\
    & \propto p(x_{1}^{T}|w_{1}^{N})p(w_{1}^{N})    
    \end{split}
\end{equation}
Therefore:
\begin{equation}
w^{*} = \operatorname{arg}\max_{w_1^N}  \underbrace{p(x_{1}^{T}|w_{1}^{N})}_{\text{acoustic model}}\cdot\underbrace{p(w_{1}^{N})}_{\text{language model}}
\end{equation}

\subsection{Node Embeddings}
Feature vectors for text in each node will be generated using selected pre-trained embedding models. Given a word sequence $w_{1}^{N}$ and a NE $k$, node features generated by embedding functions are represented as an embedding vector:
\begin{equation}
z_{1}^{d} = Embedding(w_{1}^{N})
\end{equation}

\subsection{Node Classification and Link Prediction} 
\begin{itemize}
    \item Node classification is the task of predicting the labels of nodes in a graph $G = (V, E)$ where $V$ is the set of nodes and $E$ is the set of edges. In a KG built from speech, $V$ is the set of ASR transcript $w_{1}^{N}$ and NE $k$. Given node features $Z \in \mathbb{R}^{|V| \times d}$ and (one-hot) node labels $Y \in \mathbb{R}^{|V| \times C}$ where $d$ is the number of input features associating with each node, we aim to learn a function $g: V \rightarrow \{1, \ldots, C\}$ that maps each node $v$ with its corresponding label $g(v)$.
    \item Link prediction is the task of predicting the existence of edges between node pairs in the graph. The goal is to predict a function $h: V \times V \rightarrow \{0, 1\}$ where $h(u, v) = 1$ indicates the presence of an edge between two nodes $u$ and $v$.
\end{itemize}

\subsection{GNN Models}
We explore the performance of various GNN models for node classification and link prediction. We use SAGE, GCN, GAT, and SuperGAT because they are well-suited for non-heterogeneous graphs, efficiently capturing local and global graph structure, and offering strong performance with scalable, interpretable architectures.
\begin{itemize}
    \item SAGE (Sample and Aggregate) \cite{NIPS2017_5dd9db5e}: Efficient GNN model that generates embeddings by sampling and aggregating features from a node's local neighborhood.
    \begin{equation}
    \resizebox{0.9\hsize}{!}{$
    \bm{h}_v^{(l)} = \sigma(\bm{W}_l \cdot \text{AGG}_l(\{\bm{h}_u^{(l-1)} : u \in \mathcal{N}(v)\} \cup \{\bm{h}_v^{(l-1)}\})),
    $}
    \end{equation}

    where $\mathbf{h}_v^{(l)}$ is node $v$'s hidden state at layer $l$, $\sigma$ is a non-linear activation, $\mathbf{W}_l$ is a learnable weight matrix, $\text{AGG}_l$ is an aggregation function, and $\mathcal{N}(v)$ is $v$'s neighborhood.
    
    \item GCN (Graph Convolutional Network) \cite{kipf2017semisupervised}: Spectral-based GNN learning node representations based on local graph structure.
    \begin{equation}
    \resizebox{0.75\hsize}{!}{$
    \bm{H}^{(l+1)} = \sigma(\tilde{\bm{D}}^{-\frac{1}{2}} \tilde{\bm{A}} \tilde{\bm{D}}^{-\frac{1}{2}} \bm{H}^{(l)} \bm{W}^{(l)})
    $}
    \end{equation}
    where $\mathbf{H}^{(l)}$ is the $l$-th layer activation matrix, $\tilde{\mathbf{A}} = \mathbf{A} + \mathbf{I}_N$, $\tilde{\mathbf{D}}$ is $\tilde{\mathbf{A}}$'s degree matrix, and $\mathbf{W}^{(k)}$ is a trainable weight matrix, $\mathbf{I}_N$ is is the identity matrix.
    \item GAT (Graph Attention Network) \cite{gat2018}: Assigns attention weights to neighboring nodes, focusing on the most informative ones.
    \begin{equation}
    \mathbf{h}_u' = \sigma\bigg(\sum_{v \in \mathcal{N}(u) \cup \{u\}} \alpha_{uv} \mathbf{W} \mathbf{h}_v\bigg),
    \end{equation}
    with attention coefficient defined as:
    \begin{equation}
        \alpha_{uv} = \frac{\exp(e_{uv})}{\sum_{t\in\mathcal{N}(u) \cup \{u\}} \exp(e_{ut})},
    \end{equation}
    \begin{equation}
        e_{uv} = \text{LeakyReLU}(\mathbf{a}^T [\mathbf{W} \mathbf{h}_u \| \mathbf{W} \mathbf{h}_v]),
    \end{equation}
    where $\mathbf{a}$ is a learnable attention vector.
    
    \item SuperGAT \cite{kim2021how}: Extends GAT to incorporate node and edge features in the attention mechanism.
    \begin{equation}
        \mathbf{h}_u' = \sigma(\sum_{v \in \mathcal{N}(u) \cup \{u\}} \alpha_{uv} \mathbf{W} [\mathbf{h}_v \| \mathbf{e}_{uv}]),
    \end{equation}
    where $\mathbf{e}_{uv}$ is the edge feature vector. SuperGAT uses an auxiliary loss:
    \begin{equation}
        \mathcal{L}_{att} = \sum_{u,v} \|\alpha_{uv} - \alpha_{uv}^*\|^2,
    \end{equation}
    with $\alpha_{uv}^*$ denotes the ground truth attention weight.
\end{itemize}

\section{Experimental Setups}
\label{sec:Experimental_setups}

\subsection{ASR Models}
\label{sec:ASR_models}
We employed hybrid ASR approach to transcribe audio to text. First, we generated Gaussian Mixture Model (GMM) labels as input for Deep Neural Network / Hidden Markov Model (DNN/HMM) training. We employed wav2vec 2.0 encoder \cite{baevski2020wav2vec} for the DNN, which was unsupervised pre-trained on either monolingual data or multilingual data. Their WERs on the test set were 29.0\% and 28.8\%, respectively. More details are shown in Section \ref{sec:details_ASR_models} in the Appendix.

\subsection{KG and GNN Models}
\label{sec.model_and_training}
\textbf{GNN model training}: Our research includes three training settings. Initially, the data is divided into training (60\%), validation (20\%), and testing (20\%) sets. Each GNN model is trained on the training set, with hyperparameter tuning on the validation set using metrics such as AP and AUC, then validate on test sets (known as inductive graph learning). In the second setting, models are trained on one dataset and inferenced on an unseen dataset, meaning the two KGs are independent in this scenario (known as transductive graph learning). Lastly, models are trained on the complete dataset from the first setting and directly inferenced on two KGs extracted from ASR (transductive learning) with WER 28.8\% - monolingual acoustic pre-training, and 29.0\% - multilingual acoustic pre-training. More details are shown in Section \ref{sec:details_KG_GNN_models} in the Appendix.

\noindent \textbf{Embeddings}: Our study investigates the influence of pre-trained embeddings on node classification and link prediction tasks within a KG. We used both encoder-based embeddings (such as the English Alibaba-NLP/gte-large-en-v1.5 \cite{li2023towards}, the multilingual intfloat/multilingual-e5-large-instruct \cite{wang2024multilingual}, and Vietnamese vinai/phobert-base-v2 \cite{phobert}) and decoder-based embedding (like the multilingual LLM Alibaba-NLP/gte-Qwen2-7B-instruct \cite{qwen2}), as well as random embeddings \cite{paszke2019pytorch}. Feature vectors for text in each node will be generated using selected pre-trained embedding models. More details are shown in Section \ref{sec:details_node_embeddings} in the Appendix. 

\subsection{Evaluation Metrics}
In our study, we employ two evaluation metrics: Average Precision Score (AP) and Area Under the Receiver Operating Characteristic Curve (ROC AUC or AUC) to assess the performance of our GNN models on both node classification and link prediction tasks. Details are shown in Section \ref{sec:details_evaluation_metrics} in the Appendix.

\section{Experimental Results}

\subsection{Node Classification on Human Transcript}
\subsubsection{Inductive Learning}
\begin{table}[h!]
\centering
\caption{Evaluation Results for Node Classification Task using Inductive Graph Learning on Human Transcript}

\resizebox{\columnwidth}{!}{
\begin{tabular}{llcc}
\toprule
\textbf{Model} & \textbf{Embedding} & \textbf{AP Score} & \textbf{AUC Score} \\  
\midrule
SAGE & random & 0.9116 & 0.8373 \\  
SAGE & Alibaba-NLP/gte-large-en-v1.5 & 1 & 1 \\  
SAGE & intfloat/multilingual-e5-large-instruct & 1 & 1 \\  
SAGE & vinai/phobert-base-v2 & 1 & 1 \\  
SAGE & Alibaba-NLP/gte-Qwen2-7B-instruct & 0.8017 & 0.8714 \\  \midrule
GCN & random & 0.7824 & 0.5333 \\  
GCN & Alibaba-NLP/gte-large-en-v1.5 & 0.7704 & 0.5 \\  
GCN & intfloat/multilingual-e5-large-instruct & 0.7684 & 0.5 \\  
GCN & vinai/phobert-base-v2 & 0.758 & 0.5 \\  
GCN & Alibaba-NLP/gte-Qwen2-7B-instruct & 0.2344 & 0.513 \\  \midrule
GAT & random & 0.802 & 0.5849 \\  
GAT & Alibaba-NLP/gte-large-en-v1.5 & 0.9981 & 0.9963 \\  
GAT & intfloat/multilingual-e5-large-instruct & 0.7684 & 0.5 \\ 
GAT & vinai/phobert-base-v2 & 0.9968 & 0.9944 \\
GAT & Alibaba-NLP/gte-Qwen2-7B-instruct & 0.2932 & 0.6052 \\ \midrule
SuperGAT & random & 0.803 & 0.5876 \\  
SuperGAT & Alibaba-NLP/gte-large-en-v1.5 & 0.9969 & 0.9941 \\ 
SuperGAT & intfloat/multilingual-e5-large-instruct & 0.7684 & 0.5 \\ 
SuperGAT & vinai/phobert-base-v2 & 0.9968 & 0.9941 \\  
SuperGAT & Alibaba-NLP/gte-Qwen2-7B-instruct & 0.2838 & 0.5887 \\ 
\bottomrule
\end{tabular}
\label{tab:node_evaluation_results}
}
\end{table}
The experimental results in Table \ref{tab:node_evaluation_results} show that using pre-trained embeddings significantly enhances node classification performance across all models, with perfect AP and AUC scores of 1, compared to lower scores (AP: 0.9116, AUC: 0.8373 for SAGE with random embeddings and AP: 0.8017, AUC: 0.714 for Alibaba-NLP/gte-Qwen2-7B-instruct).

Utterance nodes generally have a degree greater than 1, while named\_entity nodes have a degree of 1; the SAGE architecture uses node degrees and local neighborhood features, leading to perfect model accuracy.

The significant contribution of pre-trained embeddings and architecture to model performance is evident, with GAT and SuperGAT models achieving high AP and AUC scores using Alibaba-NLP/gte-large-en-v1.5 embeddings, while the GCN model shows limited variation in performance with different embeddings, indicating its potential limitations of leveraging these embeddings.

The Alibaba-NLP/gte-Qwen2-7B-instruct embedding exhibits poor performance across most models, indicating that high-dimensional LLM embeddings can suffer from the curse of dimensionality, and the learned information may be insufficient when there are not enough data points to establish a general pattern  as recently argued in some other natural language processing (NLP) tasks \cite{petukhova2024text, wang2023improving}.

\subsubsection{Transductive Learning}
The evaluation results for the node classification task in Table \ref{tab:discrete_node_evaluation_results} demonstrate varied performance across different models and embeddings. Among the models, SAGE combined with random embedding shows relatively high AP and AUC scores (AP: 0.792, AUC: 0.8564), indicating robust performance.
\begin{table}[h!]
\centering
\caption{Evaluation Results for Node Classification Task using Transductive Graph Learning on Human Transcript}
\resizebox{\columnwidth}{!}{
\begin{tabular}{|l|l|c|c|}
\hline
\textbf{Model} & \textbf{Embedding} & \textbf{AP Score} & \textbf{AUC Score} \\ \hline
SAGE & random & 0.792 & 0.8564 \\ \hline
SAGE & Alibaba-NLP/gte-large-en-v1.5 & 0.5326 & 0.5596 \\ \hline
SAGE & intfloat/multilingual-e5-large-instruct & 0.7009 & 0.7929 \\ \hline
SAGE & vinai/phobert-base-v2 & 0.7426 & 0.8222 \\ \hline
SAGE & Alibaba-NLP/gte-Qwen2-7B-instruct & 0.7488 & 0.8541 \\ \hline
GCN & random & 0.2983 & 0.5343 \\ \hline
GCN & Alibaba-NLP/gte-large-en-v1.5 & 0.3004 & 0.5295 \\ \hline
GCN & intfloat/multilingual-e5-large-instruct & 0.3042 & 0.532 \\ \hline
GCN & vinai/phobert-base-v2 & 0.2817 & 0.5072 \\ \hline
GCN & Alibaba-NLP/gte-Qwen2-7B-instruct & 0.2936 & 0.5214 \\ \hline
GAT & random & 0.3682 & 0.6167 \\ \hline
GAT & Alibaba-NLP/gte-large-en-v1.5 & 0.5514 & 0.5897 \\ \hline
GAT & intfloat/multilingual-e5-large-instruct & 0.3648 & 0.6145 \\ \hline
GAT & vinai/phobert-base-v2 & 0.3557 & 0.5999 \\ \hline
GAT & Alibaba-NLP/gte-Qwen2-7B-instruct & 0.3113 & 0.5468 \\ \hline
SuperGAT & random & 0.3657 & 0.6088 \\ \hline
SuperGAT & Alibaba-NLP/gte-large-en-v1.5 & 0.5488 & 0.5896 \\ \hline
SuperGAT & intfloat/multilingual-e5-large-instruct & 0.3502 & 0.5895 \\ \hline
SuperGAT & vinai/phobert-base-v2 & 0.353 & 0.5965 \\ \hline
SuperGAT & Alibaba-NLP/gte-Qwen2-7B-instruct & 0.2786 & 0.5019 \\ \hline
\end{tabular}
\label{tab:discrete_node_evaluation_results}
}
\end{table}

\subsection{Link Prediction on Human Transcript}
\subsubsection{Inductive Learning}
\begin{table}[h]
\centering
\caption{Evaluation Results for Link Prediction Task using Inductive Graph Learning on Human Transcript}
\resizebox{\columnwidth}{!}{
\begin{tabular}{llcc}
\toprule
\textbf{Model} & \textbf{Embedding} & \textbf{AP Score} & \textbf{AUC Score} \\ 
\midrule
SAGE & random & 0.4649 & 0.4912 \\  
SAGE & Alibaba-NLP/gte-large-en-v1.5 & 0.5321 & 0.5425 \\  
SAGE & intfloat/multilingual-e5-large-instruct & 0.555 & 0.593 \\  
SAGE & vinai/phobert-base-v2 & 0.7613 & 0.7869 \\  
SAGE & Alibaba-NLP/gte-Qwen2-7B-instruct & 0.4771 & 0.5219 \\  \midrule
GCN & random & 0.5263 & 0.5193 \\  
GCN & Alibaba-NLP/gte-large-en-v1.5 & 0.5504 & 0.3526 \\  
GCN & intfloat/multilingual-e5-large-instruct & 0.5 & 0.5 \\  
GCN & vinai/phobert-base-v2 & 0.6432 & 0.5284 \\  
GCN & Alibaba-NLP/gte-Qwen2-7B-instruct & 0.4934 & 0.5 \\  \midrule
GAT & random & 0.47 & 0.4617 \\  
GAT & Alibaba-NLP/gte-large-en-v1.5 & 0.7312 & 0.7242 \\  
GAT & intfloat/multilingual-e5-large-instruct & 0.5 & 0.5 \\  
GAT & vinai/phobert-base-v2 & 0.7801 & 0.8144 \\  
GAT & Alibaba-NLP/gte-Qwen2-7B-instruct & 0.5055 & 0.5102 \\  \midrule
SuperGAT & random & 0.5013 & 0.5019 \\  
SuperGAT & Alibaba-NLP/gte-large-en-v1.5 & 0.6863 & 0.681 \\  
SuperGAT & intfloat/multilingual-e5-large-instruct & 0.5 & 0.5 \\  
SuperGAT & vinai/phobert-base-v2 & 0.7522 & 0.7785 \\  
SuperGAT & Alibaba-NLP/gte-Qwen2-7B-instruct & 0.5037 & 0.5401 \\  
\bottomrule
\end{tabular}
\label{tab:link_evaluation_results}
}
\end{table}
Table \ref{tab:link_evaluation_results} shows that pre-trained embeddings greatly improve model performance in link prediction, with SAGE's AP score rising from 0.4649 to 0.7613 and its AUC score from 0.4912 to 0.7869 using vinai/phobert-base-v2 embeddings. Similarly, the GAT model's performance improves significantly with these embeddings, achieving an AP score of 0.7801 and an AUC score of 0.8144, compared to 0.47 and 0.4617 with random embeddings; however, the GCN model shows less consistent improvements. These results underscore the critical importance of embedding quality in link prediction tasks.

\subsubsection{Transductive Learning}
The link prediction task results in Table \ref{tab:discrete_link_evaluation_results} reveal differences in model effectiveness. The GCN model demonstrates superior performance, particularly when paired with random embeddings (AP: 0.9132, AUC: 0.9243) and the intfloat/multilingual-e5-large-instruct embedding (AP: 0.9015, AUC: 0.9183). In contrast, SAGE and GAT models generally exhibit lower performance, with AP and AUC scores hovering around 0.54 and 0.56, respectively. Notably, SuperGAT with intfloat/multilingual-e5-large-instruct achieves higher scores (AP: 0.6323, AUC: 0.6673), indicating a moderate level of effectiveness.
\begin{table}[h!]
\centering
\caption{Evaluation Results for Link Prediction Task using Transductive Graph Learning on Human Transcript}
\resizebox{\columnwidth}{!}{
\begin{tabular}{|l|l|c|c|}
\hline
\textbf{Model} & \textbf{Embedding} & \textbf{AP Score} & \textbf{AUC Score} \\ \hline
SAGE & random & 0.541 & 0.5615 \\ \hline
SAGE & Alibaba-NLP/gte-large-en-v1.5 & 0.5335 & 0.5429 \\ \hline
SAGE & intfloat/multilingual-e5-large-instruct & 0.5338 & 0.5497 \\ \hline
SAGE & vinai/phobert-base-v2 & 0.5123 & 0.5213 \\ \hline
SAGE & Alibaba-NLP/gte-Qwen2-7B-instruct & 0.5238 & 0.5402 \\ \hline
GCN & random & 0.9132 & 0.9243 \\ \hline
GCN & Alibaba-NLP/gte-large-en-v1.5 & 0.5968 & 0.6606 \\ \hline
GCN & intfloat/multilingual-e5-large-instruct & 0.9015 & 0.9183 \\ \hline
GCN & vinai/phobert-base-v2 & 0.8568 & 0.8515 \\ \hline
GCN & Alibaba-NLP/gte-Qwen2-7B-instruct & 0.8742 & 0.8855 \\ \hline
GAT & random & 0.5311 & 0.5576 \\ \hline
GAT & Alibaba-NLP/gte-large-en-v1.5 & 0.5419 & 0.5763 \\ \hline
GAT & intfloat/multilingual-e5-large-instruct & 0.5331 & 0.5611 \\ \hline
GAT & vinai/phobert-base-v2 & 0.5336 & 0.561 \\ \hline
GAT & Alibaba-NLP/gte-Qwen2-7B-instruct & 0.5457 & 0.5818 \\ \hline
SuperGAT & random & 0.5279 & 0.5552 \\ \hline
SuperGAT & Alibaba-NLP/gte-large-en-v1.5 & 0.5307 & 0.5583 \\ \hline
SuperGAT & intfloat/multilingual-e5-large-instruct & 0.6323 & 0.6673 \\ \hline
SuperGAT & vinai/phobert-base-v2 & 0.5086 & 0.5169 \\ \hline
SuperGAT & Alibaba-NLP/gte-Qwen2-7B-instruct & 0.556 & 0.5997 \\ \hline
\end{tabular}
\label{tab:discrete_link_evaluation_results}
}
\end{table}

\subsection{Node Classification on ASR Transcript}
\subsubsection{Monolingual Acoustic Pre-training (WER=29.0\%)}
\begin{table}[h!]
\centering
\caption{Evaluation Results for Node Classification Task on ASR Transcript using Monolingual Acoustic Pre-training}
\resizebox{\columnwidth}{!}{
\begin{tabular}{llcc}
\toprule
\textbf{Model} & \textbf{Embedding} & \textbf{AP Score} & \textbf{AUC Score} \\  
\midrule
SAGE & random & 0.8851 & 0.9204 \\  
SAGE & Alibaba-NLP/gte-large-en-v1.5 & 0.8108 & 0.8694 \\  
SAGE & intfloat/multilingual-e5-large-instruct & 0.8591 & 0.9028 \\  
SAGE & vinai/phobert-base-v2 & 0.7969 & 0.8597 \\  
SAGE & Alibaba-NLP/gte-Qwen2-7B-instruct & 0.8661 & 0.9077 \\  \midrule
GCN & random & 0.2974 & 0.5202 \\  
GCN & Alibaba-NLP/gte-large-en-v1.5 & 0.2976 & 0.521 \\  
GCN & intfloat/multilingual-e5-large-instruct & 0.2813 & 0.5052 \\  
GCN & vinai/phobert-base-v2 & 0.3037 & 0.5211 \\  
GCN & Alibaba-NLP/gte-Qwen2-7B-instruct & 0.326 & 0.5417 \\  \midrule
GAT & random & 0.2813 & 0.5808 \\  
GAT & Alibaba-NLP/gte-large-en-v1.5 & 0.3049 & 0.5406 \\  
GAT & intfloat/multilingual-e5-large-instruct & 0.3342 & 0.5758 \\  
GAT & vinai/phobert-base-v2 & 0.3334 & 0.5697 \\  
GAT & Alibaba-NLP/gte-Qwen2-7B-instruct & 0.3521 & 0.5965\\  \midrule
SuperGAT & random & 0.2778 & 0.4998 \\  
SuperGAT & Alibaba-NLP/gte-large-en-v1.5 & 0.3526 & 0.5909 \\  
SuperGAT & intfloat/multilingual-e5-large-instruct & 0.2861 & 0.5145 \\  
SuperGAT & vinai/phobert-base-v2 & 0.2751 & 0.4859 \\  
SuperGAT & Alibaba-NLP/gte-Qwen2-7B-instruct & 0.328 & 0.5682 \\  
\bottomrule
\end{tabular}
\label{tab:asr290_node_evaluation_results}
}
\end{table}
ASR-extracted utterances often lead to reduced accuracy due to noisy transcripts compared to human transcripts. The evaluation results in Table \ref{tab:asr290_node_evaluation_results} for the node classification task exhibit a consistent trend with the previous findings, but with some differences. SAGE consistently outperforms others across all embedding types, achieving the highest AP score of 0.8851 and AUC score of 0.9204 with random embeddings, surpassing even the more sophisticated embedding methods, while GCN, GAT, and SuperGAT models lag behind significantly.


\subsubsection{Multilingual Acoustic Pre-training (WER=28.8\%)}
\begin{table}[h!]
\centering
\caption{Evaluation Results for Node Classification Task with on ASR Transcript using Multilingual Acoustic Pre-training}
\resizebox{\columnwidth}{!}{
\begin{tabular}{llcc}
 \toprule
\textbf{Model} & \textbf{Embedding} & \textbf{AP Score} & \textbf{AUC Score} \\  
\midrule
SAGE & random & 0.8554 & 0.9002 \\  
SAGE & Alibaba-NLP/gte-large-en-v1.5 & 0.8307 & 0.8831 \\  
SAGE & intfloat/multilingual-e5-large-instruct & 0.8634 & 0.9058 \\  
SAGE & vinai/phobert-base-v2 & 0.8935 & 0.9266 \\  
SAGE & Alibaba-NLP/gte-Qwen2-7B-instruct & 0.9258 & 0.8959 \\  \midrule
GCN & random & 0.3015 & 0.5233 \\  
GCN & Alibaba-NLP/gte-large-en-v1.5 & 0.2811 & 0.5057 \\  
GCN & intfloat/multilingual-e5-large-instruct & 0.3215 & 0.532 \\  
GCN & vinai/phobert-base-v2 & 0.3242 & 0.5428 \\  
GCN & Alibaba-NLP/gte-Qwen2-7B-instruct & 0.7354 & 0.5325 \\  \midrule
GAT & random & 0.3315 & 0.5703 \\  
GAT & Alibaba-NLP/gte-large-en-v1.5 & 0.3052 & 0.5376 \\  
GAT & intfloat/multilingual-e5-large-instruct & 0.3609 & 0.6012 \\  
GAT & vinai/phobert-base-v2 & 0.3661 & 0.6088 \\  
GAT & Alibaba-NLP/gte-Qwen2-7B-instruct & 0.7674 & 0.6056 \\  \midrule
SuperGAT & random & 0.3519 & 0.5819 \\  
SuperGAT & Alibaba-NLP/gte-large-en-v1.5 & 0.2862 & 0.5149 \\  
SuperGAT & intfloat/multilingual-e5-large-instruct & 0.2805 & 0.5056 \\  
SuperGAT & vinai/phobert-base-v2 & 0.3268 & 0.5647 \\  
SuperGAT & Alibaba-NLP/gte-Qwen2-7B-instruct & 0.7669 & 0.6046 \\  
\bottomrule
\end{tabular}
\label{tab:asr288_node_evaluation_results}
}
\end{table}
Results in Table \ref{tab:asr288_node_evaluation_results} show patterns in model performance for the node classification task on ASR transcript using multilingual acoustic pre-training. SAGE outperforms GCN, GAT, and SuperGAT across all embedding types. Notably, SAGE combined with the vinai/phobert-base-v2 embedding achieves peak performance, with AUC score of 0.9266. SAGE is further well-performed even with random embeddings. Comparatively, multilingual acoustic pre-training shows slight improvements across most models compared to monolingual pre-training, although WERs are relatively comparable. However, the Alibaba-NLP/gte-Qwen2-7B-instruct embedding significantly outperforms other embeddings and GNN models using multilingual acoustic pre-training. Therefore, multilingual LLM text embeddings achieve optimal performance on node classification task when applied on multilingual acoustic pre-training ASR transcript.

\subsection{Link Prediction on ASR Transcript}
\subsubsection{Monolingual Acoustic Pre-training (WER=29.0\%)}
\begin{table}[h!]
\centering
\caption{Evaluation Results for Link Prediction Task on ASR Transcript using Monolingual Acoustic Pre-training}
\resizebox{\columnwidth}{!}{
\begin{tabular}{llcc}
\toprule
\textbf{Model} & \textbf{Embedding} & \textbf{AP Score} & \textbf{AUC Score} \\  
\midrule
SAGE & random & 0.5464 & 0.5701 \\  
SAGE & Alibaba-NLP/gte-large-en-v1.5 & 0.5326 & 0.5596 \\  
SAGE & intfloat/multilingual-e5-large-instruct & 0.5373 & 0.5617 \\  
SAGE & vinai/phobert-base-v2 & 0.5415 & 0.5672 \\  
SAGE & Alibaba-NLP/gte-Qwen2-7B-instruct & 0.544 & 0.5478 \\  \midrule
GCN & random & 0.8991 & 0.9259 \\  
GCN & Alibaba-NLP/gte-large-en-v1.5 & 0.9085 & 0.9262 \\  
GCN & intfloat/multilingual-e5-large-instruct & 0.9159 & 0.9324 \\  
GCN & vinai/phobert-base-v2 & 0.8811 & 0.8898 \\  
GCN & Alibaba-NLP/gte-Qwen2-7B-instruct & 0.904 & 0.9232 \\  \midrule
GAT & random & 0.5465 & 0.5825 \\  
GAT & Alibaba-NLP/gte-large-en-v1.5 & 0.5514 & 0.5897 \\  
GAT & intfloat/multilingual-e5-large-instruct & 0.5364 & 0.566 \\  
GAT & vinai/phobert-base-v2 & 0.5496 & 0.5867 \\  
GAT & Alibaba-NLP/gte-Qwen2-7B-instruct & 0.5486 & 0.5856 \\  \midrule
SuperGAT & random & 0.5601 & 0.6117 \\  
SuperGAT & Alibaba-NLP/gte-large-en-v1.5 & 0.5488 & 0.5896 \\  
SuperGAT & intfloat/multilingual-e5-large-instruct & 0.5345 & 0.5718 \\  
SuperGAT & vinai/phobert-base-v2 & 0.551 & 0.5492 \\  
SuperGAT & Alibaba-NLP/gte-Qwen2-7B-instruct & 0.5369 & 0.5677 \\  
\bottomrule
\end{tabular}
\label{tab:asr290_link_evaluation_results}
}
\end{table}
The evaluation results for the link prediction task on ASR transcript using monolingual pre-training, as presented in Table \ref{tab:asr290_link_evaluation_results}, indicate the performance of different models and embeddings. Among the models, GCN consistently outperforms SAGE, GAT, and SuperGAT in both AP and AUC scores, achieving the highest scores with the intfloat/multilingual-e5-large-instruct embedding (AP: 0.9159, AUC: 0.9324). This suggests that GCN is more effective for link prediction tasks in this context, particularly when combined with the multilingual-e5-large-instruct embedding.

\subsubsection{Multilingual Acoustic Pre-training (WER=28.8\%)}
\begin{table}[h!]
\centering
\caption{Evaluation Results for Link Prediction Task on ASR Transcript using Multilingual Acoustic Pre-training}
\resizebox{\columnwidth}{!}{
\begin{tabular}{llcc}
\toprule
\textbf{Model} & \textbf{Embedding} & \textbf{AP Score} & \textbf{AUC Score} \\  \midrule
SAGE & random & 0.551 & 0.5828 \\  
SAGE & Alibaba-NLP/gte-large-en-v1.5 & 0.5246 & 0.5532 \\  
SAGE & intfloat/multilingual-e5-large-instruct & 0.5468 & 0.5759 \\  
SAGE & vinai/phobert-base-v2 & 0.5245 & 0.5511 \\  
SAGE & Alibaba-NLP/gte-Qwen2-7B-instruct & 0.5281 & 0.5492 \\  \midrule
GCN & random & 0.8965 & 0.91 \\  
GCN & Alibaba-NLP/gte-large-en-v1.5 & 0.9184 & 0.9413 \\  
GCN & intfloat/multilingual-e5-large-instruct & 0.9 & 0.9229 \\  
GCN & vinai/phobert-base-v2 & 0.9054 & 0.9301 \\  
GCN & Alibaba-NLP/gte-Qwen2-7B-instruct & 0.869 & 0.8791 \\  \midrule
GAT & random & 0.547 & 0.5823 \\  
GAT & Alibaba-NLP/gte-large-en-v1.5 & 0.5439 & 0.5954 \\  
GAT & intfloat/multilingual-e5-large-instruct & 0.5603 & 0.6104 \\  
GAT & vinai/phobert-base-v2 & 0.5569 & 0.598 \\  
GAT & Alibaba-NLP/gte-Qwen2-7B-instruct & 0.5228 & 0.5429 \\  \midrule
SuperGAT & random & 0.5489 & 0.5867 \\  
SuperGAT & Alibaba-NLP/gte-large-en-v1.5 & 0.53 & 0.5551 \\  
SuperGAT & intfloat/multilingual-e5-large-instruct & 0.5456 & 0.5805 \\  
SuperGAT & vinai/phobert-base-v2 & 0.5336 & 0.5579 \\  
SuperGAT & Alibaba-NLP/gte-Qwen2-7B-instruct & 0.5643 & 0.6137 \\  
\bottomrule
\end{tabular}
\label{tab:asr288_link_evaluation_results}
}
\end{table}
Our empirical analysis of the link prediction task on ASR transcript using multilingual acoustic pre-training in Table \ref{tab:asr288_link_evaluation_results} also reveals a distinct performance hierarchy among GNN architectures. The GCN emerges as the preeminent model, consistently outperforming its counterparts across all embedding configurations. Notably, the GCN variant utilizing the Alibaba-NLP/gte-large-en-v1.5 embedding achieves state-of-the-art performance, with an AP score of 0.9184 and AUC score of 0.9413. This stands in stark contrast to the SAGE architecture, which, despite its prowess in node classification, exhibits suboptimal performance in this link prediction task. The GAT and SuperGAT models demonstrate intermediate efficacy, marginally surpassing SAGE but falling significantly short of GCN's benchmark. The intfloat/multilingual-e5-large-instruct embedding consistently augments model performance, but with varying magnitudes of impact across architectures.

\subsection{Error Analysis}
\textbf{Node classification and link prediction on human transcript}: In inductive learning, BERT-based embeddings are essential for achieving optimal performance, whereas in transductive learning, random text embeddings demonstrate competitiveness with pre-trained embeddings.

\noindent \textbf{Node classification on ASR transcript}: Firstly, in the context of noisy ASR transcripts, both monolingual and multilingual acoustic pre-training settings demonstrate that random text embeddings perform competitively with BERT-based text embeddings. For comparison, this transductive learning approach for ASR transcripts is similar to the transductive learning used for node classification on human transcripts. Secondly, multilingual LLM text embeddings notably outperform others in node classification tasks when applied to multilingual acoustic pre-training ASR transcripts. As our study is the first evaluation of training KGs from speech, there is no existing literature for direct comparison. However, combination of multilingual LLM text embeddings and multilingual acoustic pre-training  generally yield higher accuracy across various downstream tasks, e.g. ASR \cite{lam2023multilingual, radford2023robust}, speech translation \cite{bapna2022mslam, babu2022xls, zhang2023google}, text-to-speech \cite{saeki2023virtuoso, zhang2019learning}. Thirdly, within the same transductive learning setting, node classification on ASR transcripts generally achieved competitive results compared to human transcripts, despite relatively high WERs of 28.8\% and 29\%.

\noindent \textbf{Link prediction on ASR transcript}: Firstly, in the context of noisy ASR transcripts, both in monolingual and multilingual acoustic pre-training ASR settings, random text embeddings demonstrate performance comparable to BERT-based or LLM text embeddings. This transductive learning performance is also observed in the transductive learning of link prediction on human transcripts. Secondly, in the same transductive learning setting, link prediction on ASR transcripts generally outperformed that on human transcripts, despite relatively high WERs of 28.8\% and 29\%. This result is noteworthy, as high WERs in ASR transcripts typically degrade accuracy in various downstream NLP tasks, as widely demonstrated by the AI community \cite{desot2019slu, sundararaman2021phoneme, omachi2021end}. We hypothesized that the influence of text embedding, which primarily focuses on the generalized context of text segments (semantics), reduces the impact of ASR errors on prediction performance \cite{voleti2019investigating}. Thirdly, our transductive learning on ASR transcripts for both node classification and link prediction tasks was conducted in a zero-shot setting. We hypothesize that the adaptation of trained GNN models to the ASR transcripts of the training set could further enhance performance \cite{dinh2021zero, ma2023can}.

\section{Conclusion}
In this study, we introduce \textit{wav2graph}, the first framework for supervised learning of KG from speech data. Additionally, we present the first real-world KG derived from speech, along with its baseline results.

\noindent Our study demonstrates that, first of all, for node classification and link prediction tasks on ASR transcripts, both monolingual and multilingual acoustic pre-training with random text embeddings perform competitively with encoder-based and decoder-based embeddings. This trend is also observed in transductive learning on human transcripts. Secondly, multilingual LLM text embeddings significantly outperform other embeddings in node classification tasks when applied to multilingual acoustic pre-trained ASR transcripts. Thirdly, node classification on ASR transcripts generally achieves competitive results compared to human transcripts, while link prediction on ASR transcripts generally outperforms that on human transcripts, despite relatively high WERs of 28.8\% and 29\%. This unexpected behavior is likely due to the influence of text embeddings, which primarily focus on the generalized semantic context of text segments and therefore reduce the impact of ASR errors on prediction performance. This contrasts with most previous works on other downstream tasks, where high WERs in ASR transcripts typically degrade accuracy.

\section{Acknowledgement}
We extend our gratitude to Oanh Tran at VNU-HCM University of Technology for her assistance in preparing the paper draft.

\bibliography{custom}

\clearpage 


\appendix

\onecolumn
\tableofcontents
\newpage

\section{Related Works}
This section describes works relevant to our \textit{wav2graph} framework.

\textbf{KG from speech}: In the domain of knowledge graph construction, traditional methodologies have predominantly focused on extracting information from textual sources \cite{zhong2023comprehensive, wang2021structure, wang2014knowledge, chen2021zero}. Although there has been progress incorporating multimodal inputs, such as images and text \cite{zhu2022multi, li2024graphadapter, alberts2021visualsem, yu2021ernie}, the challenge of directly constructing knowledge graphs from speech data remains largely unexplored. Among the limited number of relevant works we could find to our best knowledge, \citet{fu2021speech} introduced what they claimed to be the first automatic KG construction system from speech. Also, \citet{wu2022towards} proposed a novel information extraction task, termed speech relation extraction, which utilized extracted relations from synthetic ASR transcripts to construct a KG. However, to the best of our knowledge, no training has been conducted on such KGs. Training KGs is crucial as GNN models can learn to extract and generalize complex patterns and relationships from the data within a KG, thereby enabling predictions on unseen data in the KG—capabilities that rule-based methods alone cannot achieve.

\textbf{Information extraction from speech}: The challenge of accurately identifying entities and their relationships within vast and unstructured data sources persists as a major barrier to broader text-based KG adoption \cite{peng2023knowledge}. Such relationships are typically extracted through NER systems \cite{li2022unified, li2020survey}. However, performing NER on speech, which is necessary for constructing voice-based KGs, remains a significant challenge \cite{chen2022aishell, sui2021large, szymanski2023aren, yadav2020end, caubriere2020we}.

\textbf{GNNs for speech applications}: GNNs have shown promise in various speech-related applications. For instance, \citet{wang2020speaker, singh2023supervised} applied GNNs to improve speaker diarization, while \citet{pentari2024speech, joshi2022cogmen, li2023speech} employed them for speech emotion recognition. Also, GNNs could be used in speaker verification task \cite{jung2021graph, he2024study, jung2022aasist} and ASR hypothesis decoding \cite{dighe2020lattice, sun2022tree}. Despite these advancements, existing GNN-based approaches do not address the direct construction and training of KGs from speech data, which enables the prediction of attributes and relations between spoken utterances. 

\textbf{GNNs}: Graph representation learning has evolved significantly in recent years, with various approaches designed to incorporate graph structure into meaningful node features. For example, Graph Convolutional Networks (GCN) \cite{kipf2016semi} and Message Passing Neural Networks (MPNN) \cite{10.5555/3305381.3305512} pioneered this field by proposing the message passing scheme, in which each node aggregates features from the adjacent nodes. Subsequent works like Graph Attention Networks (GAT) \cite{velickovic2017graph} introduced mechanisms to prioritize important nodes. However, these classical approaches often struggle with capturing long-range relationships \cite{dwivedi2022long}. To address this limitation, researchers have explored virtual nodes \cite{pham2017graph, pmlr-v202-cai23b} and k-hop neighborhoods \cite{nikolentzos2020k} within the message passing framework. More recently, graph transformers have gained prominence, with models such as TokenGT \cite{10.5555/3600270.3601330} and Graphormer \cite{ying2021transformers} incorporating sophisticated encodings such as centrality and spatial information. GraphGPS further advanced this approach by combining various positional and structural encodings with multiple graph block types \cite{rampavsek2022recipe}. Additionally, \citet{ngo2023multiresolution} proposed a multiscale graph transformer that learns hierarchical graph coarsening and utilizes graph wavelet transforms for positional encoding. These advancements in graph representation learning have significantly improved the ability to capture both local and global graph structures, paving the way for more effective graph-based machine learning models.

\section{Details about Experimental Setups}
This section describes details about experimental setups for experimental reproducibility, which extends the Section \ref{sec:Experimental_setups} in the main paper.

\subsection{ASR Models}
\label{sec:details_ASR_models}
This section extends details of Section \ref{sec:ASR_models} in the main paper.

\noindent \textbf{Gaussian Mixture Model / Hidden Markov Model (GMM/HMM)}: For language modelling and initial GMM/HMM, we followed the same setups and hyperparameters as in \cite{luscher2023development}. First, we used the toolkit Sequitur Grapheme-To-Phoneme (g2p) \cite{G2P_toolkit} to convert pronunciation lexica found in human transcript, so that the seed lexicon was extended, creating the lexica for training.  Secondly, we created an n-gram language model using previously extended lexica and human transcript based on Kneser-Ney Smoothing \cite{ney1994structuring}. Thirdly, we created alignments obtained by using GMM/HMM as labels for wav2vec 2.0 \cite{baevski2020wav2vec} neural network training, which later resulted to Deep Neural Network / Hidden Markov Model (DNN/HMM) training. The acoustic modeling employed context-dependent phonemic labels, triphones to be specific. In GMM/HMM process, we used a CART (Classification And Regression Tree) \cite{breiman2017classification} to tie the states $s$, resulting 4501 CART labels. During GMM/HMM process, we stopped at Speaker Adaptive Training stage (SAT) \cite{miao2015speaker} instead of going beyond Speaker Adaptive Training + Vocal Tract Length Normalization \cite{eide1996parametric} (SAT+VTLN) for the sake of good WER labels: 
\begin{equation}
\begin{split}
&p(x_1^T|w_1^N) = \sum_{[s_1^T]}\prod_{t=1}^Tp(x_t, s_t|s_{t-1}, w_1^N) \\
&= \sum_{[s_1^T]}\prod_{t=1}^T\underbrace{p(s_t|s_{t-1}, w_1^N)}_{\text{transition prob.}}\cdot \underbrace{p(x_t|s_t, s_{t-1}, w_1^N)}_{\text{emission prob.}}    
\end{split}
\end{equation}

\noindent \textbf{Unsupervised DNN pre-training}: For unsupervised DNN pre-training, we used wav2vec 2.0 \cite{facebook2020wav2vec2} as DNN encoder with the help of Fairseq \cite{facebook2019fairseq} toolkit. We employed similar vanilla configurations and hyperparameters as in \cite{bachelorthesis}. All models had 118M parameters including 7 Convolutional Neural Network (CNN) \cite{fukushima1979neural, fukushima1980neocognitron} layers and 8 Transformer \cite{vaswani2017attention} layers. The last CNN layer had a stride halved for the 8kHz data \cite{vieting2023efficient}.

\noindent \textbf{DNN/HMM training}: We loaded unsupervised pre-trained wav2vec 2.0 models for fine-tuning in a supervised DNN/HMM approach. We optimized the model with Framewise Cross-Entropy (fCE) loss. The SpecAugment \cite{park2019specaugment} data augmentation was applied for the entire 33 fine-tuning epochs. We used RETURNN toolkit \cite{zeyer2018returnn} for supervised training. 

\noindent \textbf{ASR hypothesis decoding}: The entire ASR hypothesis decoding was done using RASR toolkit \cite{rybach2011rasr}. In this stage, we integrated the acoustic model with the n-gram language model using Bayes' decision rule and the Viterbi algorithm \cite{viterbi}. The Viterbi algorithm recursively computes the optimal path through the alignment graph of all potential word sequences, thereby identifying the best alignment with the acoustic observations. Then, pruning of the acoustic model and the n-gram language model through beam search was employed to focus exclusively on the most likely predicted words at each time step $t$ \cite{ortmanns1997word}. Viterbi algorithm is described as:
\begin{equation}
\begin{split}
w_1^N &= \operatorname{arg}\max_{N,w_1^N}p\Bigl(\prod_{n=1}^Np(w_n|w_{n-m}^{n-1}) \\
&\cdot \max_{[s_1^T]}\prod_{t=1}^Tp(x_t,s_t|s_{t-1}, w_1^N)\Bigr)    
\end{split}
\end{equation}

\noindent \textbf{Monolingual and multilingual pre-training}: We utilized two best baseline models for ASR task on the \textit{VietMed} dataset, as described by \citet{vietmed_dataset}. The models employed were a monolingual acoustic pre-trained w2v2-Viet and a multilingual acoustic pre-trained XLSR-53-Viet. The w2v2-Viet model was pre-trained on 1204 hours of unlabeled Vietnamese data, whereas the XLSR-53-Viet model was pre-trained on 1204 hours of unlabeled Vietnamese data with an initialization from the multilingual pre-trained XLSR-53 \cite{conneau21_XLSR53}. Both models possess  118M parameters and were fine-tuned using the same training set. Their WERs on the test set were 29.0\% and 28.8\%, respectively.

\subsection{KG and GNN Models}
\label{sec:details_KG_GNN_models}
This section extends details of Section \ref{sec.model_and_training} in the main paper.

\noindent \textbf{KG preprocessing}: The KG is preprocessed, involving the identification of entity and relation types, attribute normalization, and potential feature engineering for edge features. The KG consists of two types of nodes: utterance (e.g., "Doctors go to hospital") and named\_entity (e.g., "doctors"(PERSON) and "hospital" (LOCATION)). NEs are extracted from the utterances.

\noindent \textbf{Hyperparameter Tuning}: Hyperparameter tuning will focus on optimizing the combination of hidden layers and message passing aggregation functions for each GNN model. In the first setting, models will be trained with fixed hyperparameters: 250 epochs (10 epochs for SAGE), a learning rate of 0.005, weight decay of 0.05, Adam optimizer \cite{kingma2014adam}, and dropout rates \cite{srivastava2014dropout} of 0.2 for node classification tasks and 0.5 for link prediction tasks. The same hyperparameters will be applied for both the second and third settings.

\subsection{Node Embeddings}
\label{sec:details_node_embeddings}
This section extends details of Section \ref{sec.model_and_training} in the main paper. 

We utilized both encoder-based embeddings, including the English Alibaba-NLP/gte-large-en-v1.5 \cite{li2023towards}, the multilingual intfloat/multilingual-e5-large-instruct \cite{wang2024multilingual}, and the Vietnamese vinai/phobert-base-v2 \cite{phobert}, as well as decoder-based embeddings, such as the multilingual LLM Alibaba-NLP/gte-Qwen2-7B-instruct \cite{qwen2}, in addition to random embeddings \cite{paszke2019pytorch}. We used the following models to get embeddings:
\begin{itemize}
    \item Random embedding: Features are initialized randomly\footnote{https://pytorch.org/docs/stable/generated/torch.randn.html} \cite{paszke2019pytorch}, a tensor filled with random numbers from a normal distribution with mean 0 and variance 1, serving as a baseline for embedding comparison.
    \item Alibaba-NLP/gte-large-en-v1.5 \cite{li2023towards}: General-purpose general text embeddings with multi-stage contrastive learning that built upon the encoder backbone (BERT \cite{devlin2019bert} + RoPE \cite{su2024roformer} + GLU \cite{shazeer2020glu}).
    \item intfloat/multilingual-e5-large-instruct \cite{wang2024multilingual}: The opensource multilingual E5 text embedding models, released in mid-2023. The training procedure adheres to the English E5 model recipe, involving contrastive pre-training on 1 billion multilingual text pairs, followed by fine-tuning on a combination of labeled datasets.
    \item vinai/phobert-base-v2 \cite{phobert}: A pre-trained RoBERTa \cite{liu2019roberta} language models for Vietnamese. 
    \item Alibaba-NLP/gte-Qwen2-7B-instruct \cite{qwen2}: It is the latest model in the gte (General Text Embedding \cite{li2023towards}) model family with 7 billion parameters.
\end{itemize}

\subsection{Details about Evaluation Metrics}
\label{sec:details_evaluation_metrics}

\textbf{AP metric}: The AP summarizes the precision-recall curve as the weighted mean of precisions achieved at each threshold, with the increase in recall from the previous threshold used as the weight. It provides a single number to summarize the classifier's performance, which is especially useful in the context of imbalanced datasets where one class may be underrepresented. For the node classification task, let $\hat{c}_v$ be the true label of node $v$ and  $c_v$ be the predicted probability of node $v$ belonging to a specific class. For the link prediction task, let $\hat{e}_{u,v}$ be the true label indicating the presence of an edge between nodes $u$ and $v$, and ${e}_{u,v}$ be the predicted probability of an edge existing between nodes $u$ and $v$. We then sort the nodes in descending order of their predicted probabilities $c_v$ and ${e}_{u,v}$ respectively. Finally, the AP score is calculated as:
\begin{equation}
    \text{AP} = \sum_n (R_n - R_{n-1}) P_n,
\end{equation}
where $P_n$ and $R_n$ are the precision and recall at the $n$-th threshold, respectively.

\noindent \textbf{AUC metric}: AUC represents the degree or measure of separability, indicating how much the model is capable of distinguishing between classes. The ROC curve is a graphical plot that illustrates the diagnostic ability of a binary classifier system as its discrimination threshold is varied. $TPR(t)$ and $FPR(t)$ be the true positive rate and false positive rate at threshold $t$, respectively. The AUC is computed as:
\begin{equation}
    \text{AUC} = \int_{0}^{1} TPR(FPR^{-1}(x)) \, dx.
\end{equation}
This integral calculates the area under the ROC curve.

\onecolumn

\section{Additional Experimental Results}
This section shows the cross-validation loss curves over steps for all GNN models (SAGE, GCN, GAT, and SuperGAT) and all 5 embeddings. 

\subsection{Node Classification on Human Transcript}
This section shows the cross-validation loss curves of node classification task on human transcript, which are derived from Table \ref{tab:node_evaluation_results} in the main paper.

\begin{figure}[h]
    \centering
    \includegraphics[width=0.7\linewidth]{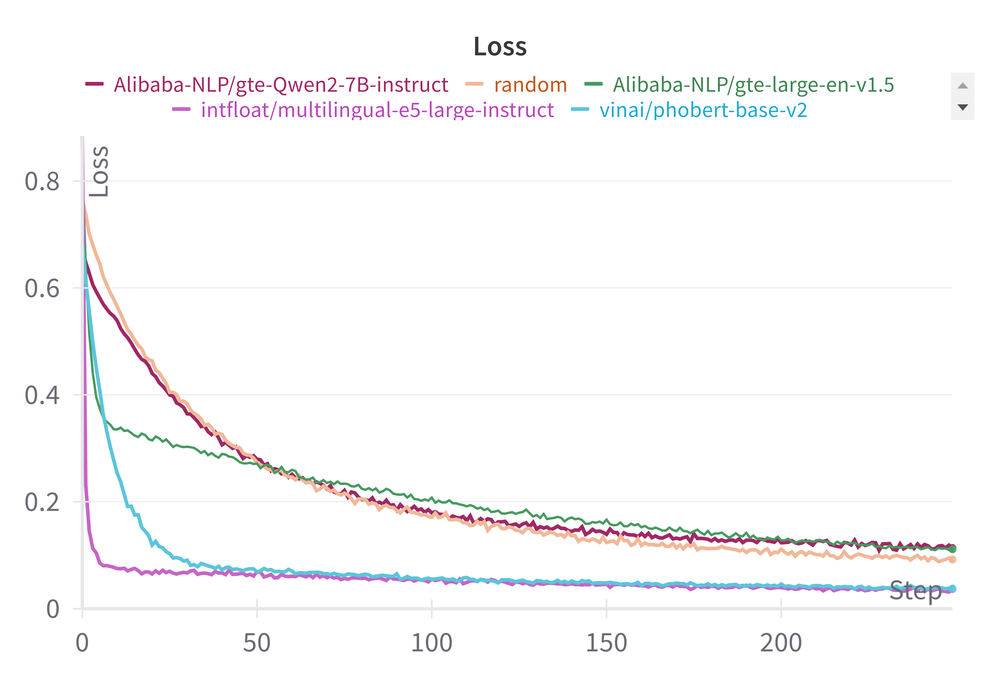}
    \caption{Loss at each iteration with SAGE model.}
    \label{fig:enter-label}
\end{figure}

\begin{figure}[h]
    \centering
    \includegraphics[width=0.7\linewidth]{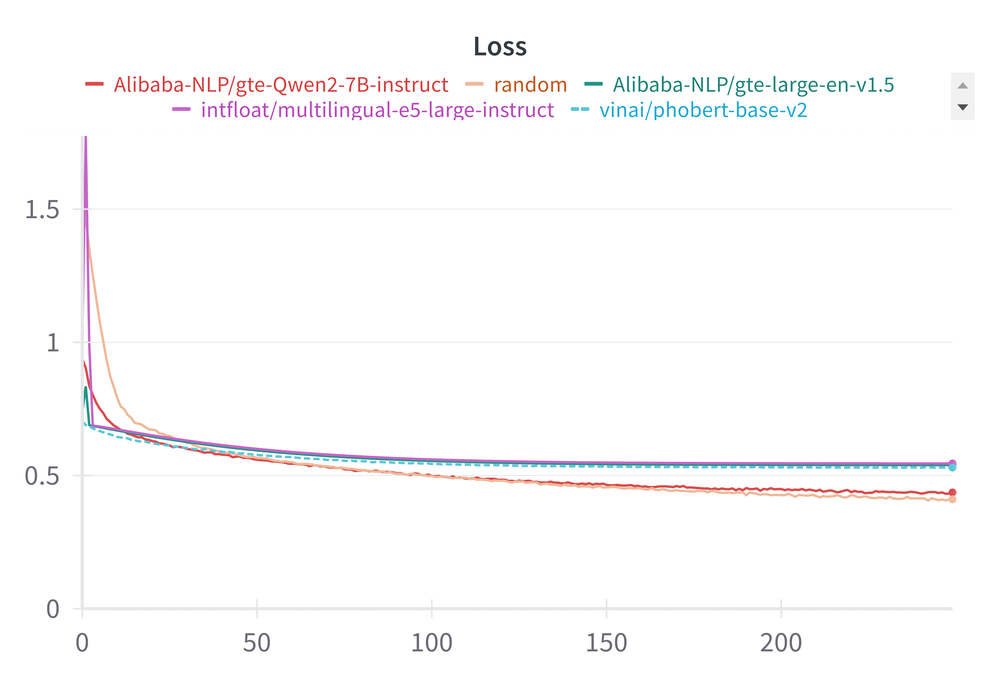}
    \caption{Loss at each iteration with GCN model.}
    \label{fig:enter-label}
\end{figure}

\begin{figure}[h]
    \centering
    \includegraphics[width=0.7\linewidth]{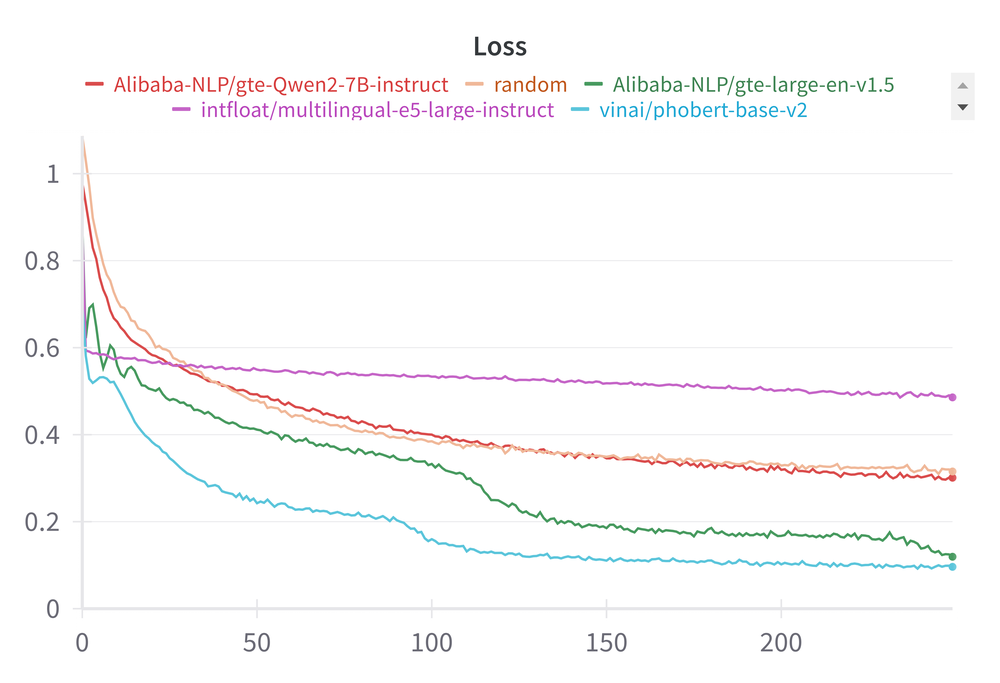}
    \caption{Loss at each iteration with GAT model.}
    \label{fig:enter-label}
\end{figure}

\begin{figure}[h]
    \centering
    \includegraphics[width=0.7\linewidth]{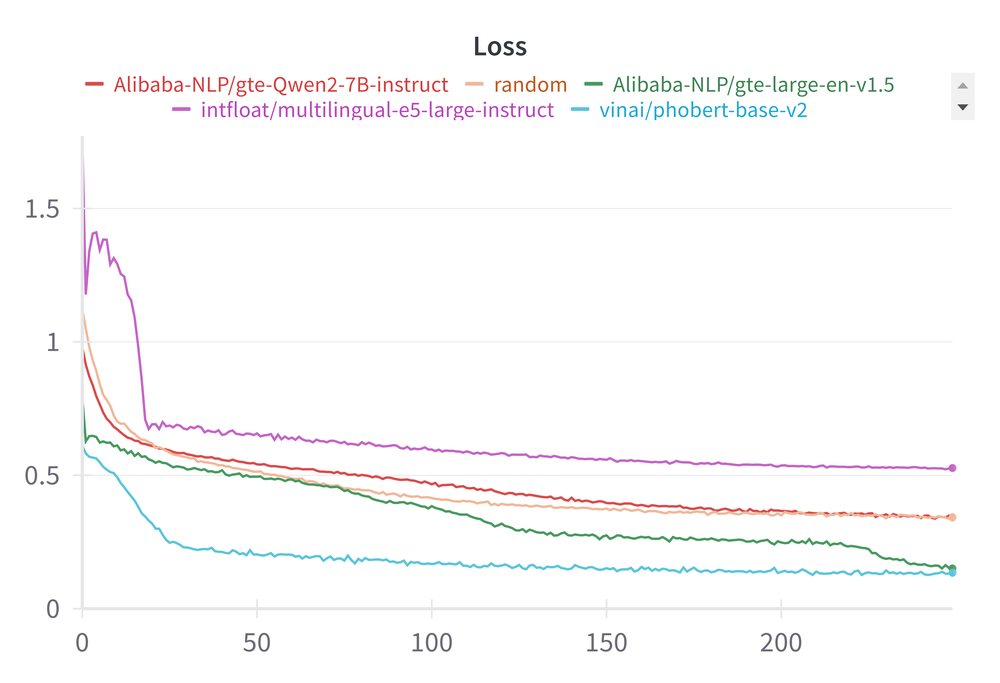}
    \caption{Loss at each iteration with SuperGAT model.}
    \label{fig:enter-label}
\end{figure}

\onecolumn

\subsection{Link Prediction on Human Transcript}
This section shows the cross-validation loss curves of link prediction task on human transcript, which are derived from Table \ref{tab:link_evaluation_results} in the main paper.

\begin{figure}[h]
    \centering
    \includegraphics[width=0.7\linewidth]{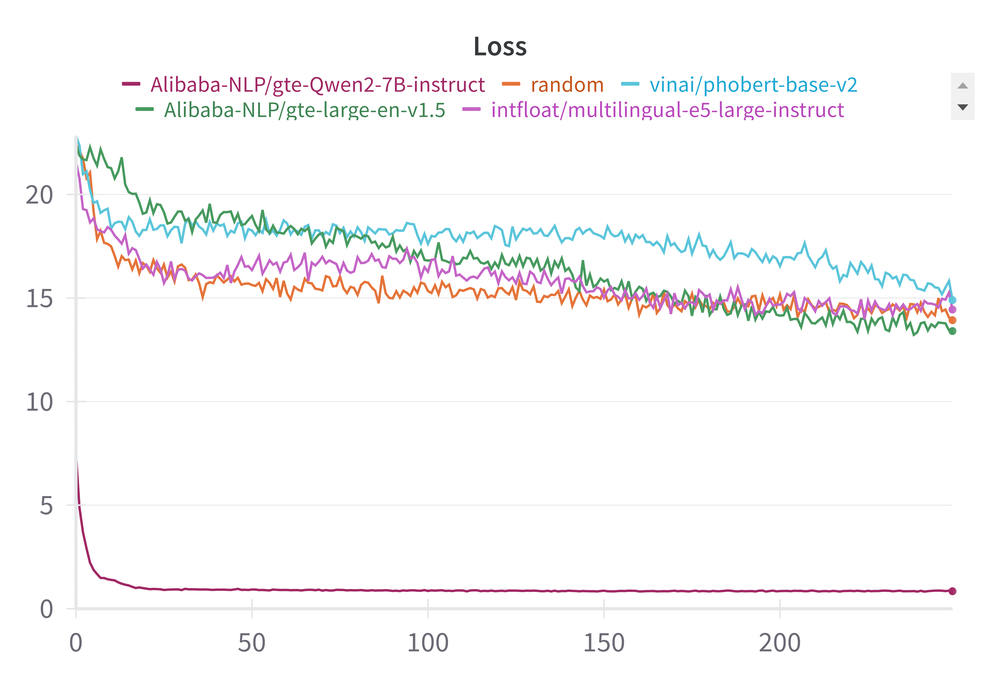}
    \caption{Loss at each iteration with SAGE model.}
    \label{fig:enter-label}
\end{figure}

\begin{figure}[h]
    \centering
    \includegraphics[width=0.7\linewidth]{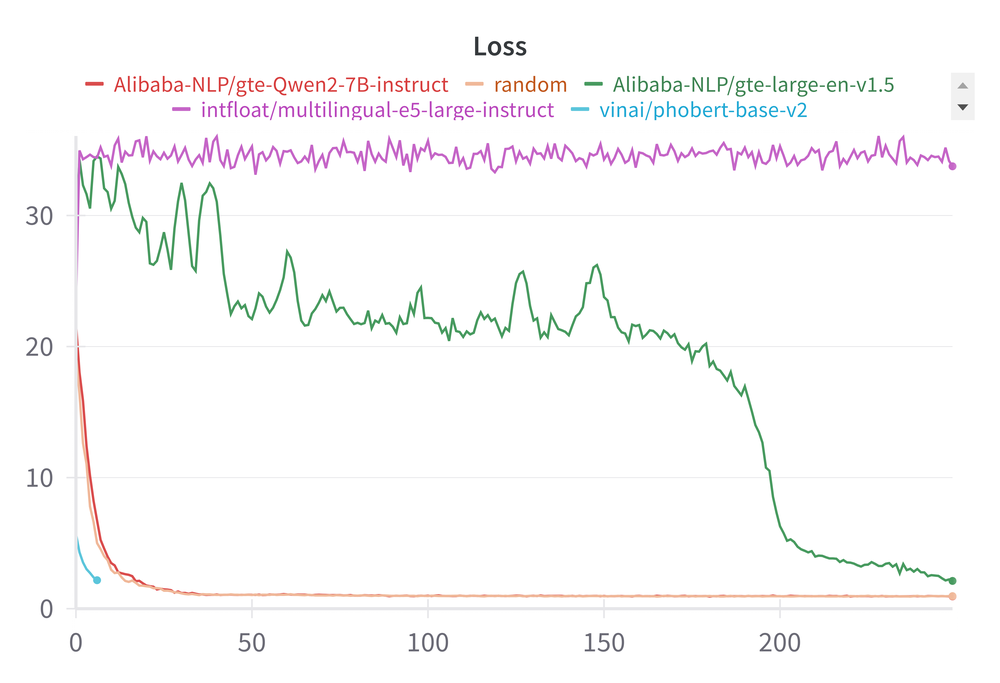}
    \caption{Loss at each iteration with GCN model.}
    \label{fig:enter-label}
\end{figure}

\begin{figure}[h]
    \centering
    \includegraphics[width=0.7\linewidth]{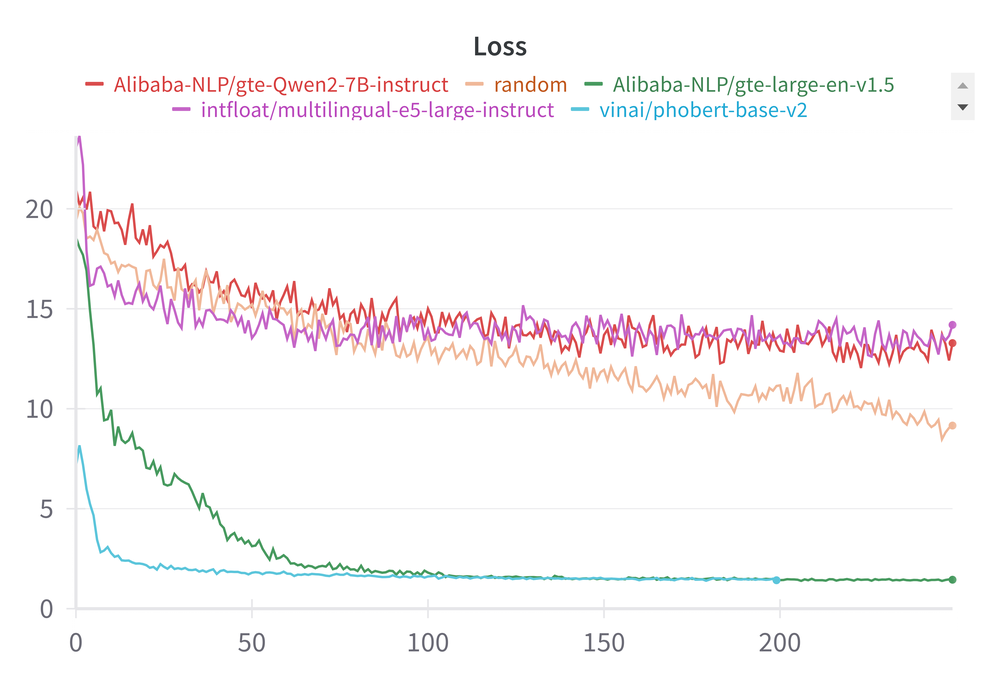}
    \caption{Loss at each iteration with GAT model.}
    \label{fig:enter-label}
\end{figure}

\begin{figure}[h]
    \centering
    \includegraphics[width=0.7\linewidth]{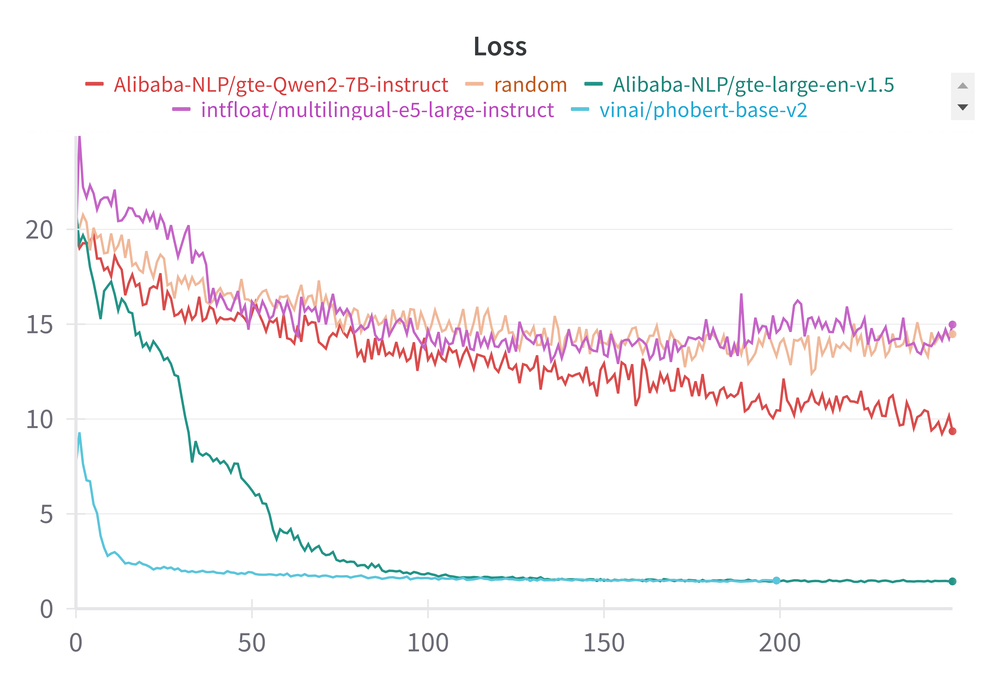}
    \caption{Loss at each iteration with SuperGAT model.}
    \label{fig:enter-label}
\end{figure}

\onecolumn

\subsection{Node Classification on ASR Transcript}

\subsubsection{Monolingual Acoustic Pre-training (WER=29.0\%)}
This section shows the cross-validation loss curves of node classification task on ASR transcript using monolingual acoustic pre-training, which are derived from Table \ref{tab:asr290_node_evaluation_results} in the main paper.

\begin{figure}[h]
    \centering
    \includegraphics[width=0.7\linewidth]{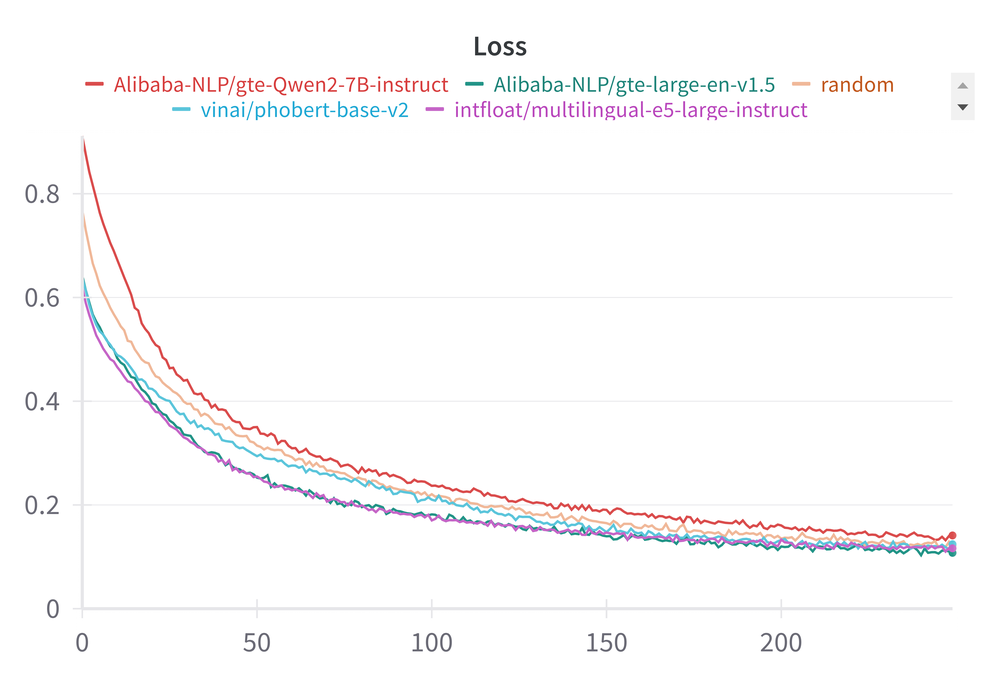}
    \caption{Loss at each iteration with SAGE model.}
    \label{fig:enter-label}
\end{figure}

\begin{figure}[h]
    \centering
    \includegraphics[width=0.7\linewidth]{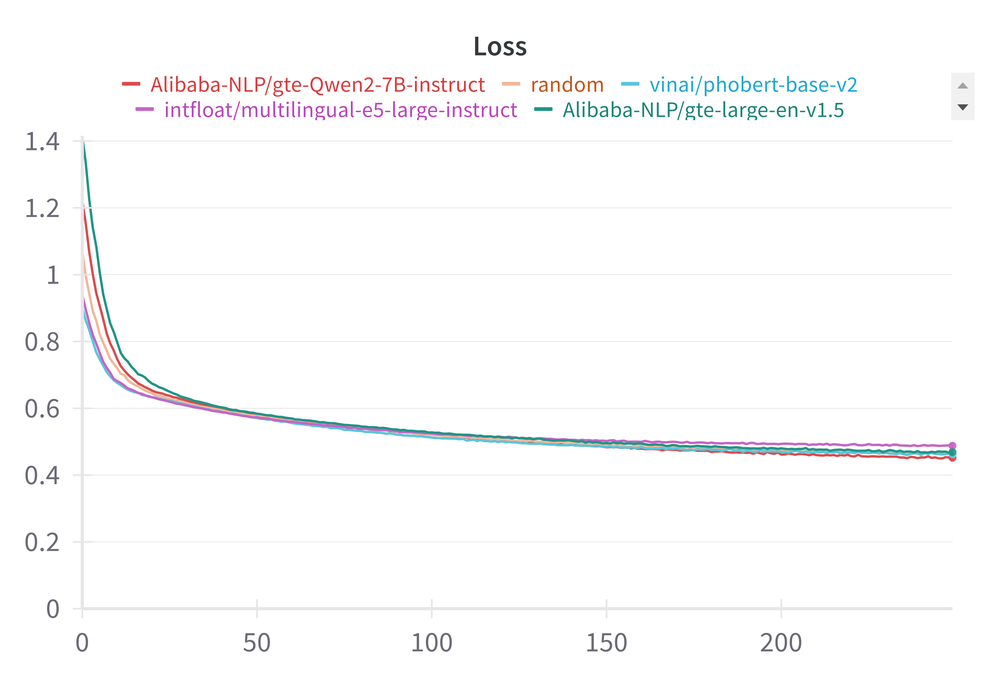}
    \caption{Loss at each iteration with GCN model.}
    \label{fig:enter-label}
\end{figure}

\begin{figure}[h]
    \centering
    \includegraphics[width=0.7\linewidth]{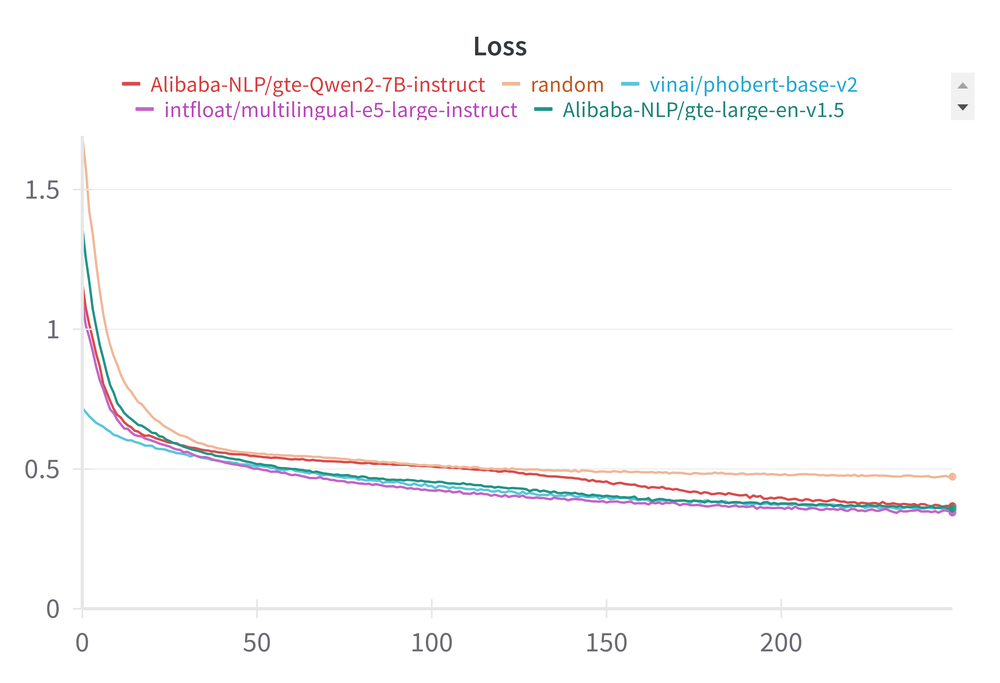}
    \caption{Loss at each iteration with GAT model.}
    \label{fig:enter-label}
\end{figure}

\begin{figure}[h]
    \centering
    \includegraphics[width=0.7\linewidth]{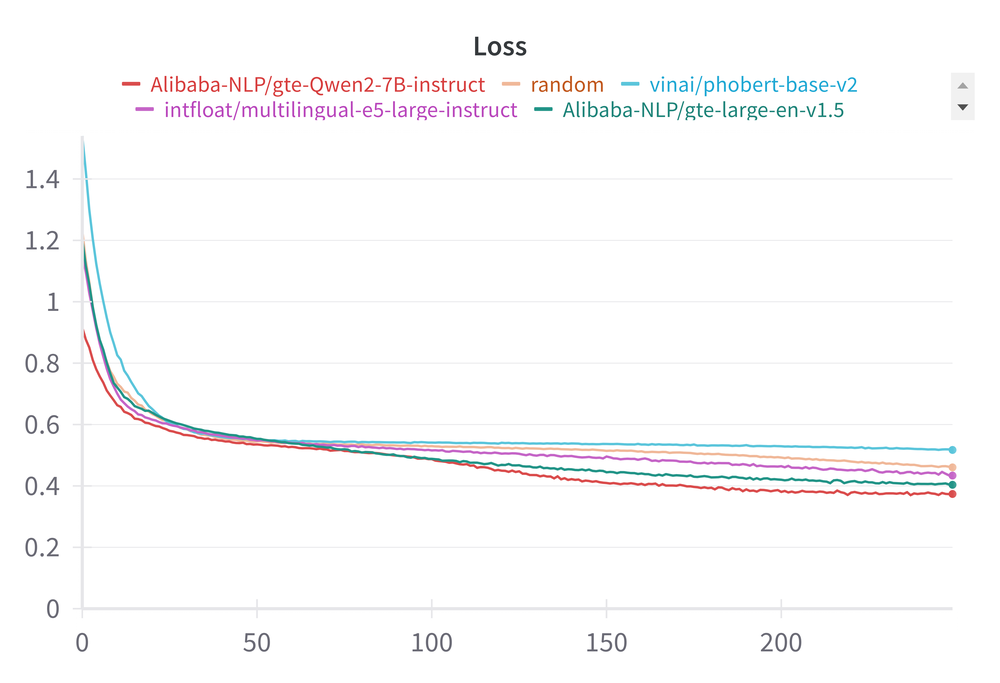}
    \caption{Loss at each iteration with SuperGAT model.}
    \label{fig:enter-label}
\end{figure}

\onecolumn

\subsubsection{Multilingual Acoustic Pre-training (WER=28.8\%)}
This section shows the cross-validation loss curves of node classification task on ASR transcript using multilingual acoustic pre-training, which are derived from Table \ref{tab:asr288_node_evaluation_results} in the main paper.

\begin{figure}[h]
    \centering
    \includegraphics[width=0.7\linewidth]{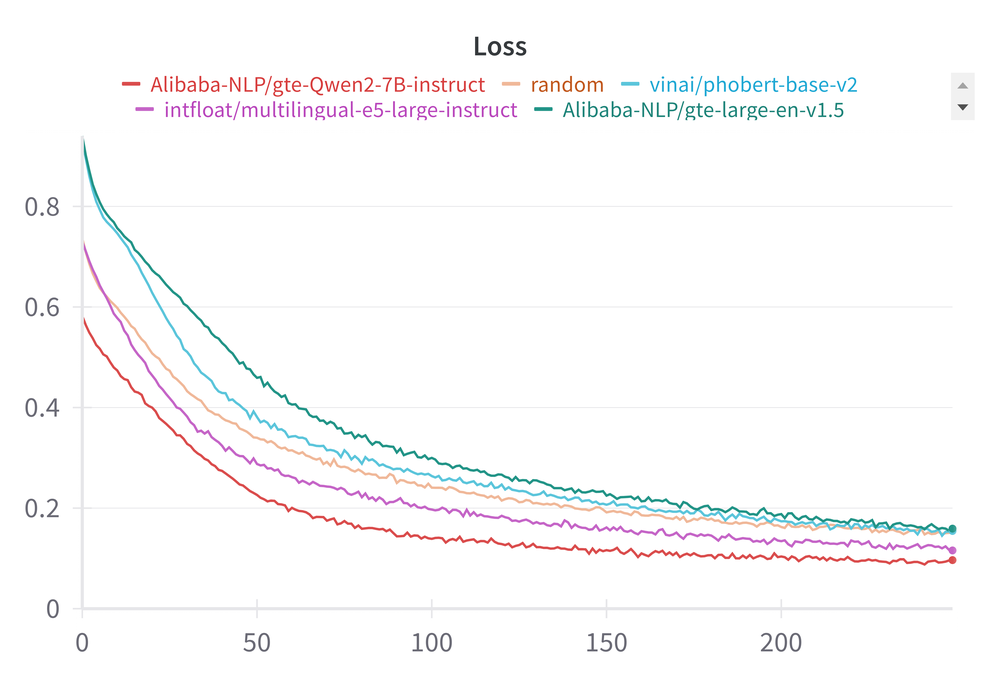}
    \caption{Loss at each iteration with SAGE model.}
    \label{fig:enter-label}
\end{figure}

\begin{figure}[h]
    \centering
    \includegraphics[width=0.7\linewidth]{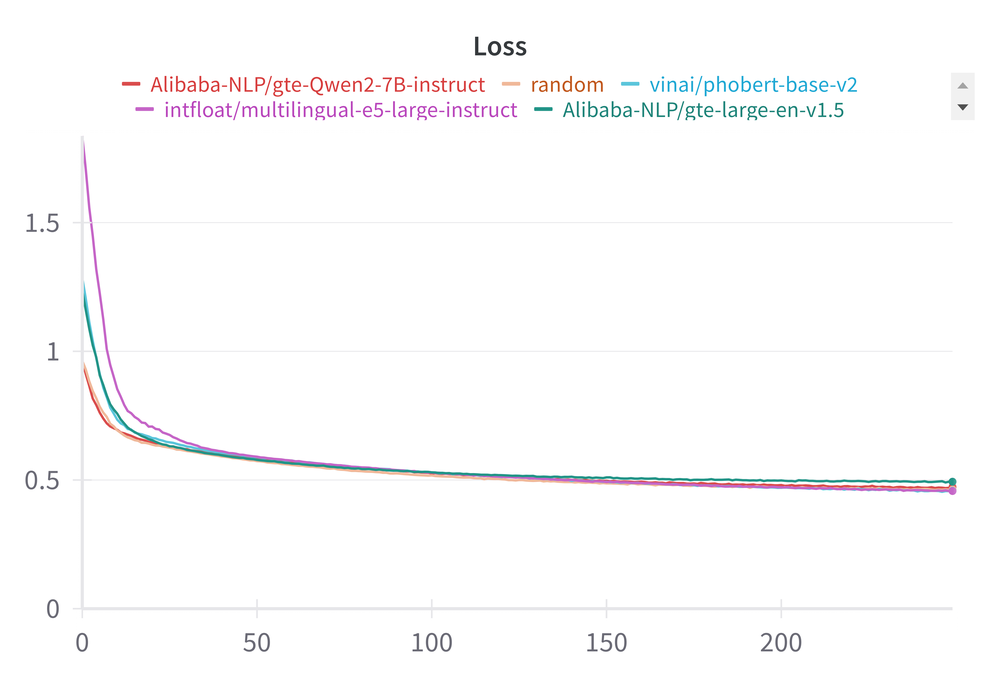}
    \caption{Loss at each iteration with GCN model.}
    \label{fig:enter-label}
\end{figure}

\begin{figure}[h]
    \centering
    \includegraphics[width=0.7\linewidth]{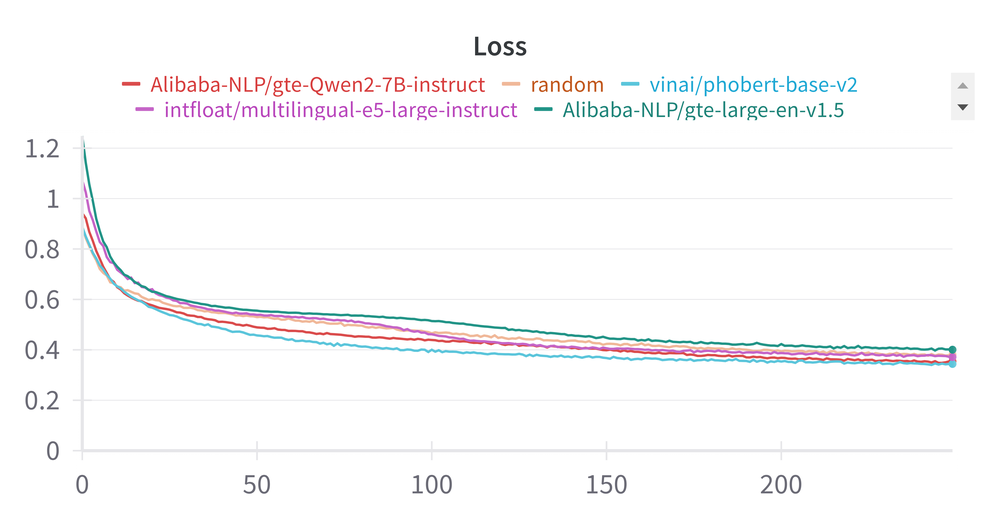}
    \caption{Loss at each iteration with GAT model.}
    \label{fig:enter-label}
\end{figure}

\begin{figure}[h]
    \centering
    \includegraphics[width=0.7\linewidth]{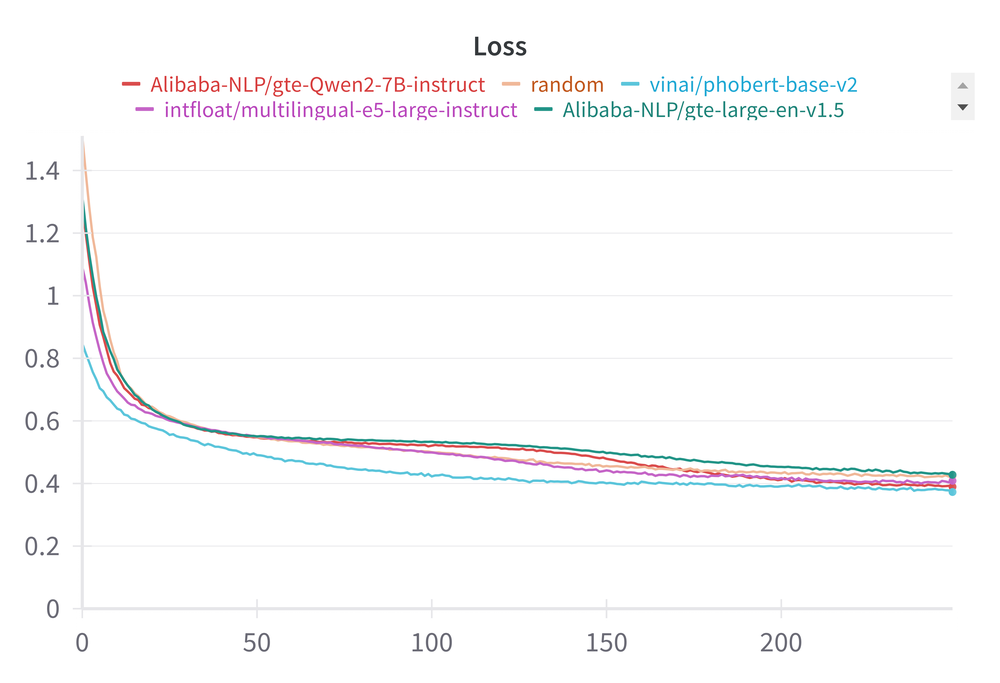}
    \caption{Loss at each iteration with SuperGAT model.}
    \label{fig:enter-label}
\end{figure}

\onecolumn

\subsection{Link Prediction on ASR Transcript}

\subsubsection{Monolingual Acoustic Pre-training (WER=29.0\%)}
This section shows the cross-validation loss curves of link prediction task on ASR transcript using monolingual acoustic pre-training, which are derived from Table \ref{tab:asr290_link_evaluation_results} in the main paper.

\begin{figure}[h]
    \centering
    \includegraphics[width=0.7\linewidth]{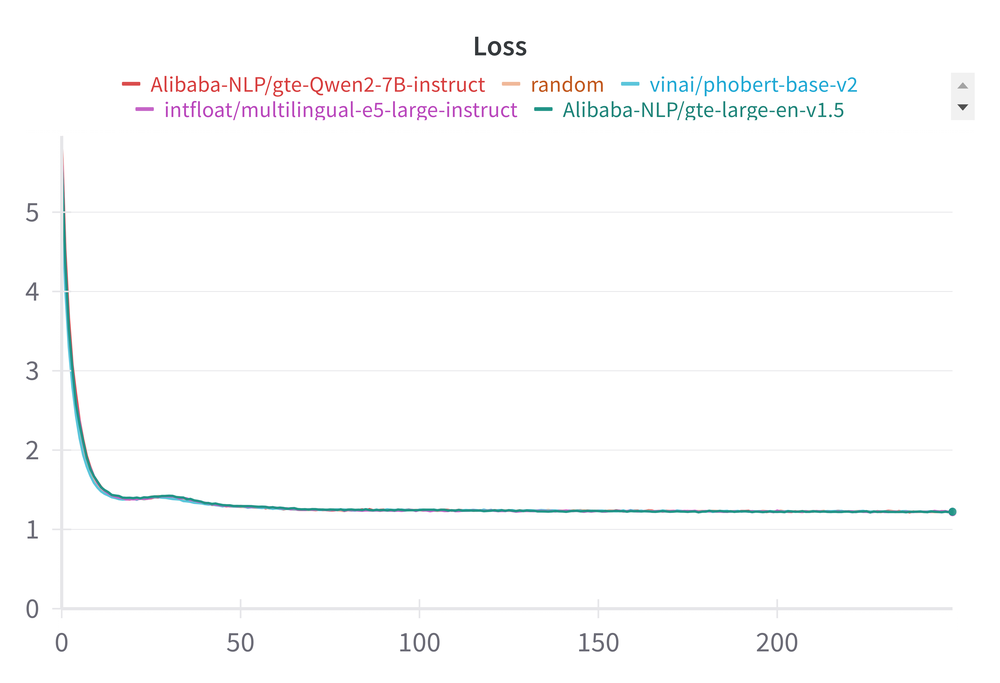}
    \caption{Loss at each iteration with SAGE model.}
    \label{fig:enter-label}
\end{figure}

\begin{figure}[h]
    \centering
    \includegraphics[width=0.7\linewidth]{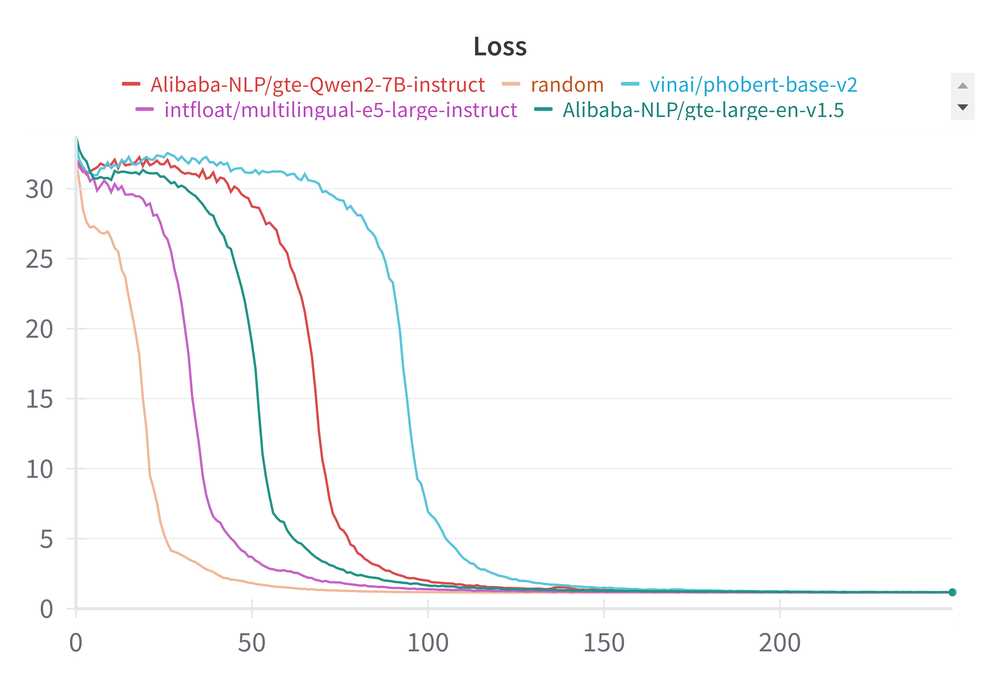}
    \caption{Loss at each iteration with GCN model.}
    \label{fig:enter-label}
\end{figure}

\begin{figure}[h]
    \centering
    \includegraphics[width=0.7\linewidth]{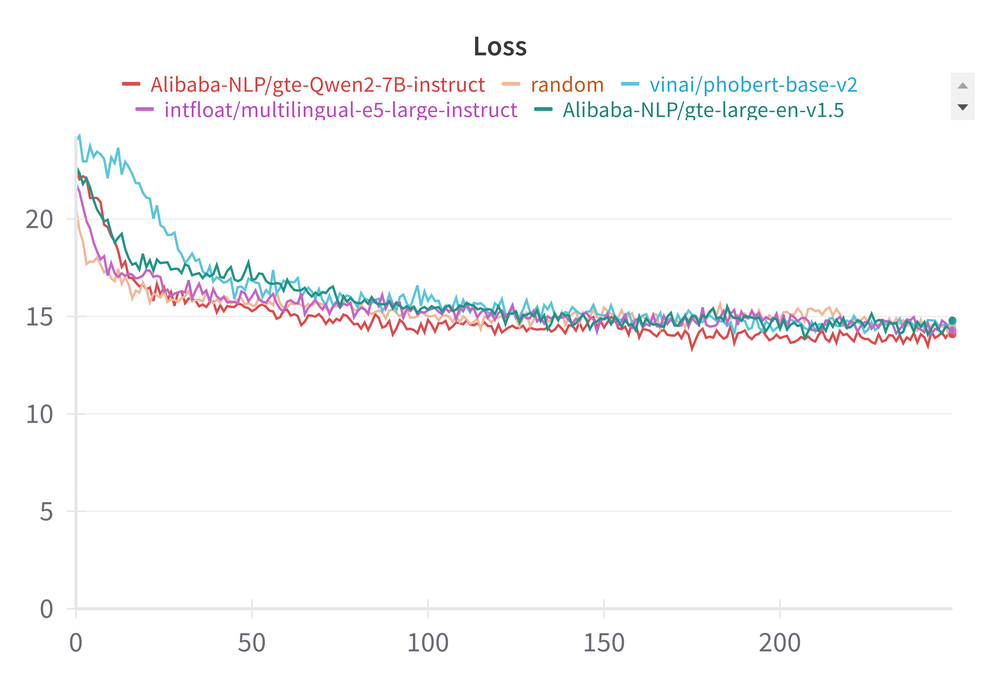}
    \caption{Loss at each iteration with GAT model.}
    \label{fig:enter-label}
\end{figure}

\begin{figure}[h]
    \centering
    \includegraphics[width=0.7\linewidth]{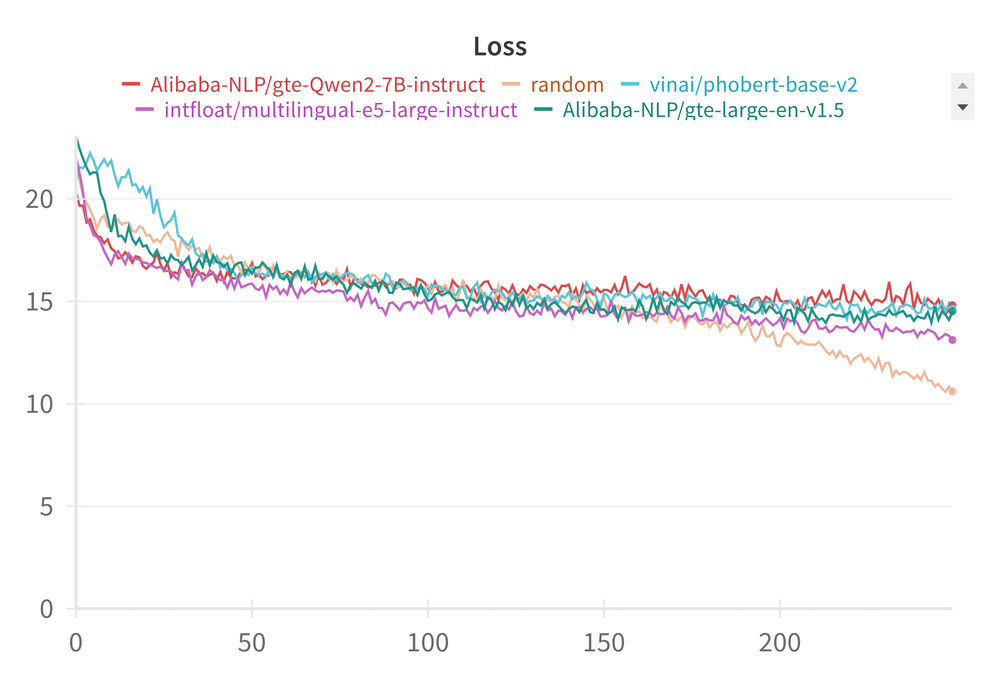}
    \caption{Loss at each iteration with SuperGAT model.}
    \label{fig:enter-label}
\end{figure}

\onecolumn

\subsubsection{Multilingual Acoustic Pre-training (WER=28.8\%)}
This section shows the cross-validation loss curves of link prediction task on ASR transcript using multilingual acoustic pre-training, which are derived from Table \ref{tab:asr288_link_evaluation_results} in the main paper.

\begin{figure}[h]
    \centering
    \includegraphics[width=0.7\linewidth]{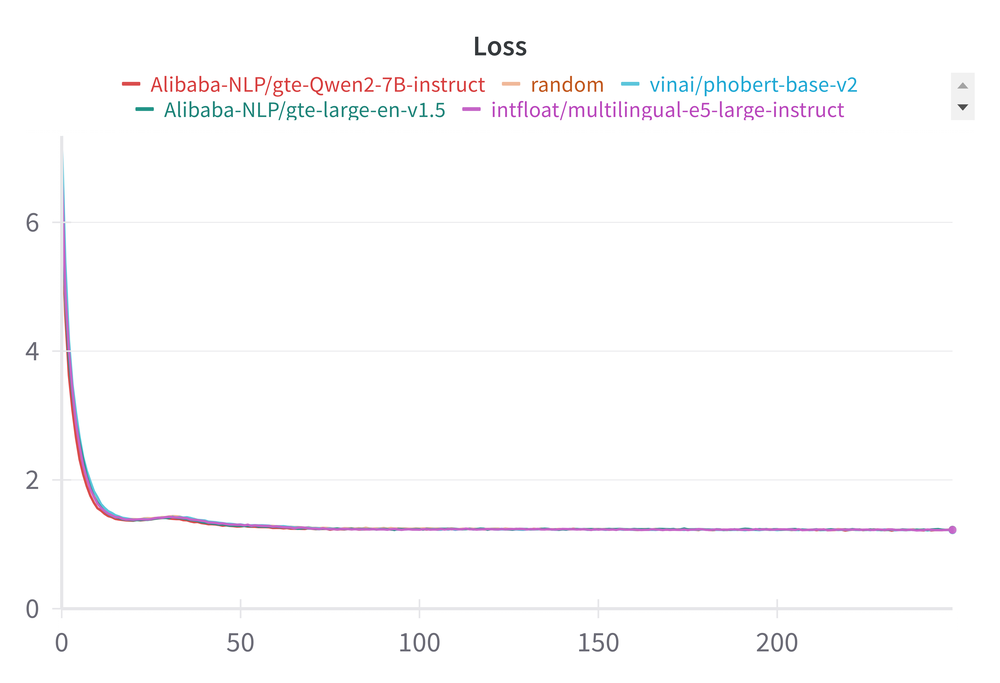}
    \caption{Loss at each iteration with SAGE model.}
    \label{fig:enter-label}
\end{figure}

\begin{figure}[h]
    \centering
    \includegraphics[width=0.7\linewidth]{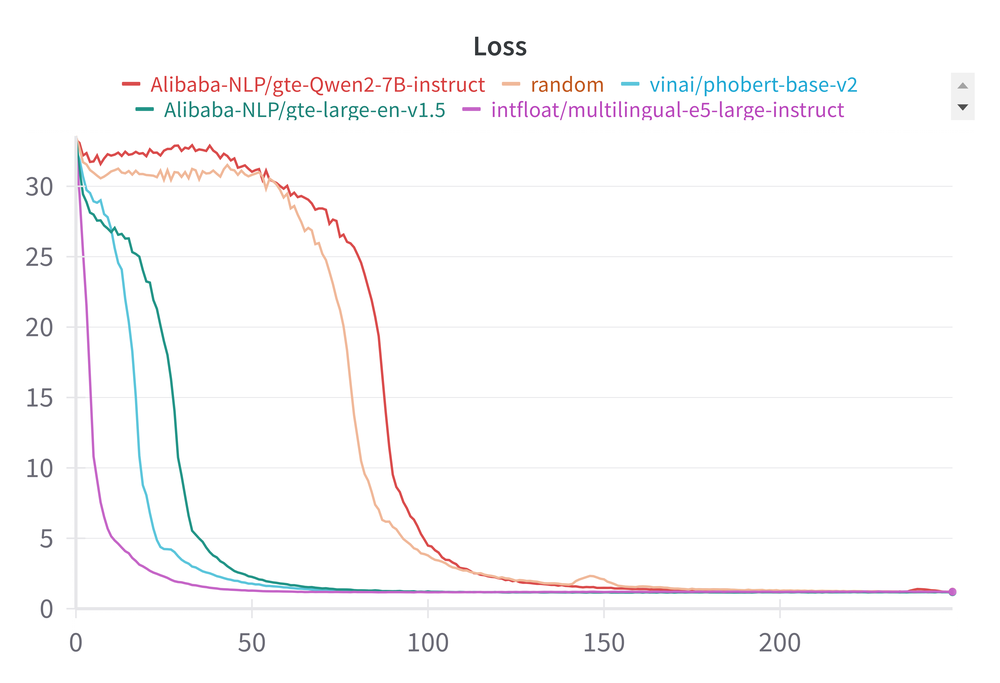}
    \caption{Loss at each iteration with GCN model.}
    \label{fig:enter-label}
\end{figure}

\begin{figure}[h]
    \centering
    \includegraphics[width=0.7\linewidth]{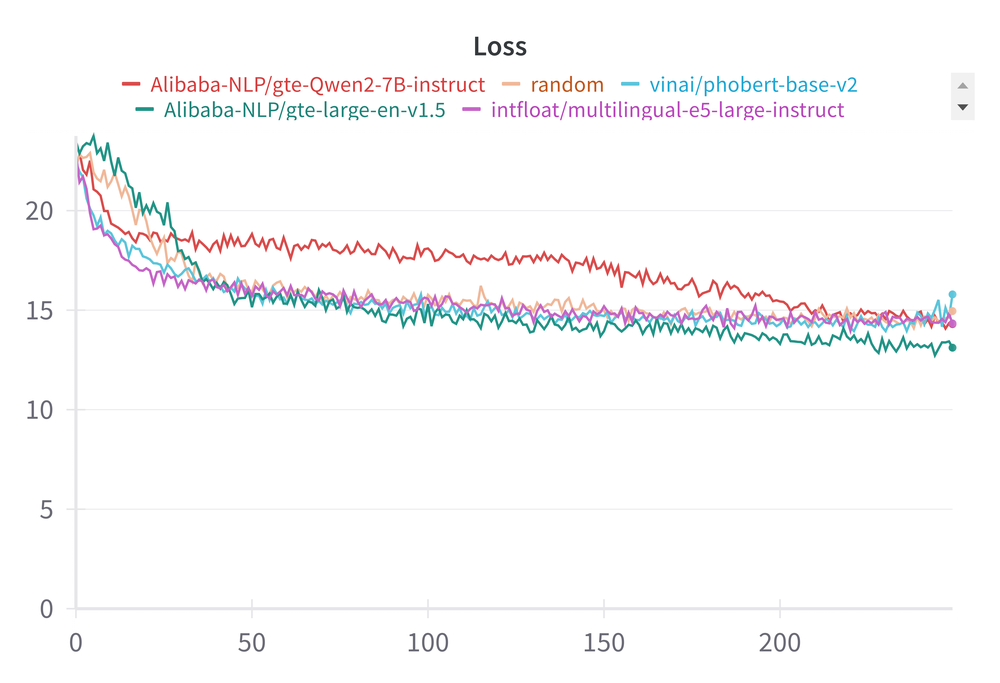}
    \caption{Loss at each iteration with GAT model.}
    \label{fig:enter-label}
\end{figure}

\begin{figure}[h]
    \centering
    \includegraphics[width=0.7\linewidth]{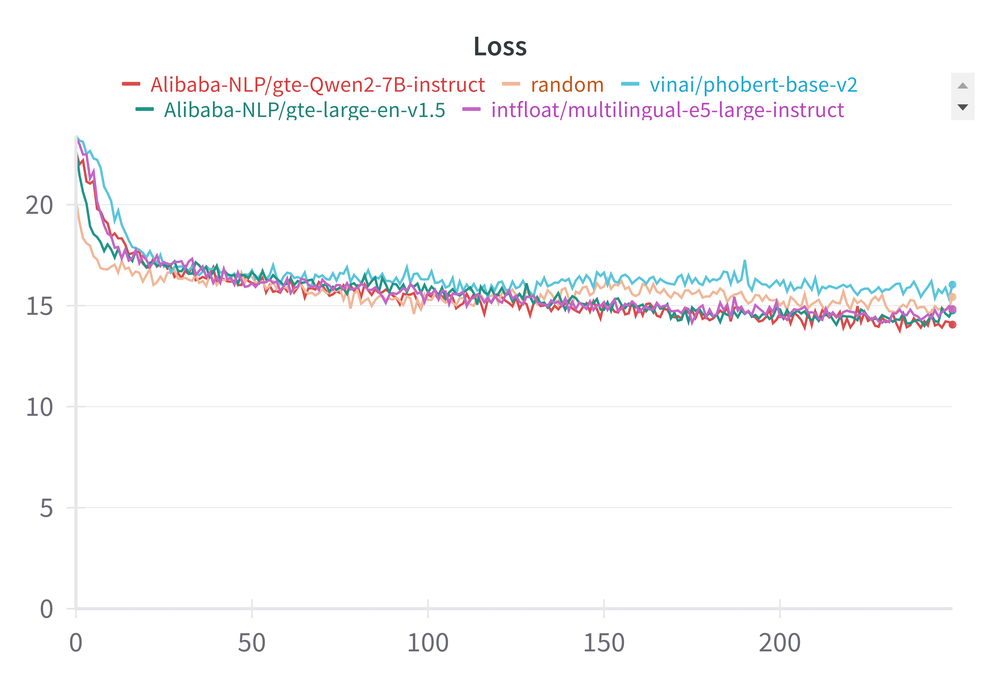}
    \caption{Loss at each iteration with SuperGAT model.}
    \label{fig:enter-label}
\end{figure}

\end{document}